\documentclass[10pt]{article}

\usepackage{geometry}
\usepackage{srcltx,graphicx}
\usepackage[export]{adjustbox}   

\usepackage{amsmath,amssymb,amsthm,amscd,mathrsfs,bm}
\usepackage[unicode,colorlinks,linkcolor=blue,hyperindex]{hyperref}
\usepackage{indentfirst} 
\usepackage{enumerate} 
\usepackage{cases} 
\usepackage[mathscr]{euscript}
\usepackage{wasysym}
\usepackage{algorithm}
\usepackage{algpseudocode}
\usepackage{tikz-cd}
\usepackage{arydshln}
\usepackage{colortbl}
\usepackage{multirow}
\usepackage{subcaption} 
\usepackage{booktabs}%
\usepackage{tikz}
\usetikzlibrary{shapes.geometric}
\usetikzlibrary{arrows.meta,calc}
\usetikzlibrary{positioning}
\usepackage{xcolor} 

\graphicspath{{figure/}}

\usepackage{lineno}
\makeatletter
\renewcommand{\p@subfigure}{\thefigure}
\makeatother
\graphicspath{figure}
\numberwithin{equation}{section} \numberwithin{figure}{section}

{\theoremstyle{remark} }

\numberwithin{figure}{section}
\numberwithin{table}{section}
\newcommand{\normm}[1]{{\left\vert\kern-0.25ex\left\vert\kern-0.25ex\left\vert #1 
		\right\vert\kern-0.25ex\right\vert\kern-0.25ex\right\vert}}

\begin{document} 
\title{Learning Patient-Specific Spatial Biomarker Dynamics via Operator Learning for Alzheimer’s Disease Progression}
\author{Jindong Wang\textsuperscript{a}, Yutong Mao\textsuperscript{b}, Xiao Liu\textsuperscript{b},
	Wenrui Hao\textsuperscript{a}\textsuperscript{*},\\   {for the Alzheimer's Disease Neuroimaging Initiative$^\dagger$}\\
}

\date{}
\maketitle
\vspace{-3em} 
\begin{center}\scriptsize
	\textsuperscript{a}\textit{Department of Mathematics, Penn State University, University Park, PA, USA}\\
    	\textsuperscript{b}\textit{Department of Biomedical Engineering, Penn State University, University Park, PA, USA}\\
        
	\textsuperscript{*}\textit{Corresponding author}: \texttt{wxh64@psu.edu}\\
    \textsuperscript{$\dagger$}\textit{A list of authors and their affiliations appears at the end of the paper. }
\end{center}
\begin{abstract}
Alzheimer’s disease (AD) is a complex, multifactorial neurodegenerative disorder with substantial heterogeneity in progression and treatment response. Despite recent therapeutic advances, predictive models capable of accurately forecasting individualized disease trajectories remain limited. Here, we present a machine learning–based operator learning framework for personalized modeling of AD progression, integrating longitudinal multimodal imaging, biomarker, and clinical data. Unlike conventional models with prespecified dynamics, our approach directly learns patient-specific disease operators governing the spatiotemporal evolution of amyloid, tau, and neurodegeneration biomarkers. Using Laplacian eigenfunction bases, we construct geometry-aware neural operators capable of capturing complex brain dynamics. Embedded within a digital twin paradigm, the framework enables individualized predictions, simulation of therapeutic interventions, and in silico clinical trials. Applied to AD clinical data, our method achieves high prediction accuracy exceeding 90\% across multiple biomarkers, substantially outperforming existing approaches. This work offers a scalable, interpretable platform for precision modeling and personalized therapeutic optimization in neurodegenerative diseases.
\end{abstract}
\section{Introduction}

Alzheimer’s disease (AD) remains one of the most pressing global health challenges, affecting over 50 million people worldwide. Despite recent FDA approvals of anti-amyloid therapies such as Aducanumab (2021) and Lecanemab (2023), their clinical benefits, particularly in slowing cognitive decline, remain limited and controversial~\cite{pmid36847013,pmid36872303}. With more than 140 therapeutics currently in clinical trials and significant investment of resources, the persistently high failure rate underscores fundamental gaps in our understanding of AD’s complex, multifactorial pathophysiology and the substantial heterogeneity in disease trajectories and treatment responses~\cite{pmid36818565,pmid36804755,pmid37183523}.

The classical pathological hallmarks of AD, amyloid-$\beta$ plaques and tau neurofibrillary tangles, remain central to diagnosis and therapeutic targeting, yet their precise roles in driving cognitive decline are incompletely understood~\cite{pmid36834612,pmid36829875}. Mounting evidence characterizes AD as a heterogeneous disorder involving genetic, vascular, metabolic, and neuroinflammatory factors~\cite{pmid36807325,michno2022chemical}. As a result, population-based models and one-size-fits-all treatment approaches often fail to capture patient-specific disease mechanisms and therapeutic responses, posing a major obstacle for precision medicine in AD.

Computational modeling presents a promising approach to addressing this challenge by simulating disease dynamics, elucidating causal interactions, and evaluating personalized therapeutic interventions in silico. Several computational models have been developed to characterize AD biomarker progression, including mechanistic ordinary differential equation (ODE) models~\cite{pmid30863455,petrella2019computational,rabiei2025data}, statistical latent time models~\cite{raket2020statistical,li2018bayesian}, and disease progression scoring frameworks~\cite{gueorguieva2023disease,iturria2016early}. More recently, systems biology approaches have integrated omics and imaging data to infer causal relationships among biomarkers~\cite{pmid20083042,iturria2017multifactorial}. While these models have advanced our understanding, most rely on prespecified model forms and fixed assumptions, limiting their capacity to capture the diverse, nonlinear progression patterns observed in individual patients. A central open question remains: {\em how can data itself reveal the governing dynamical models of disease progression?}

To address this, growing attention has turned to the concept of \textit{digital twins} — computational replicas of individual patients constructed from multimodal clinical, imaging, and biomarker data to simulate disease trajectories and predict therapeutic outcomes. However, a critical bottleneck persists: how to efficiently and interpretably learn personalized disease models directly from high-dimensional, multimodal patient data. This motivates the integration of machine learning with operator-based modeling, a methodology well-established in fields such as fluid dynamics and materials science, but largely unexplored in neurodegenerative disease modeling.

In this study, we introduce a novel \textit{machine learning–based operator learning framework} for modeling AD progression. Unlike conventional machine learning models that map finite-dimensional inputs to outputs, operator learning methods are designed to learn mappings between function spaces, enabling the direct approximation of the dynamical operators that govern disease biomarker and imaging trajectories over time. This eliminates the need for prespecified model structures, allowing disease progression dynamics to emerge directly from longitudinal data.

Our framework further leverages the \textit{digital twin paradigm} to build individualized computational models using multimodal patient data. By learning patient-specific disease operators, we can generate personalized predictions of future disease trajectories, simulate therapeutic interventions, and conduct virtual clinical trials entirely in silico. Importantly, the framework is generalizable: once the disease operator is learned, it can be used to simulate different therapeutic scenarios, optimize treatment regimens, and forecast personalized outcomes, all within a mechanistically interpretable and data-informed modeling environment.

Technically, we employ \textit{Laplacian eigenfunction bases} to parameterize neural operators, providing an efficient, geometry-aware spectral representation well-suited to structured and unstructured brain domains. This spectral formulation enables robust learning of spatiotemporal dynamics of amyloid, tau, neurodegeneration, and cognitive decline from longitudinal imaging and biomarker data within the Alzheimer’s Disease Neuroimaging Initiative (ADNI) cohort. Beyond predictive accuracy, the learned operators offer mechanistic insight through the analysis of operator Jacobians and spatial interaction patterns.

Overall, this work establishes a scalable, interpretable, and personalized computational framework for operator learning in neurodegenerative disease modeling. By integrating data-driven operator learning with the digital twin paradigm, our approach represents an important step toward precision modeling of complex, heterogeneous diseases like AD, with direct applications to virtual clinical trials, individualized therapeutic optimization, and mechanistic disease insight.

\section{Results}

\subsection{Learning Spatiotemporal PDE Models from Imaging Data}

\begin{figure}[!htbp]
    \centering
    \includegraphics[width=.9\linewidth]{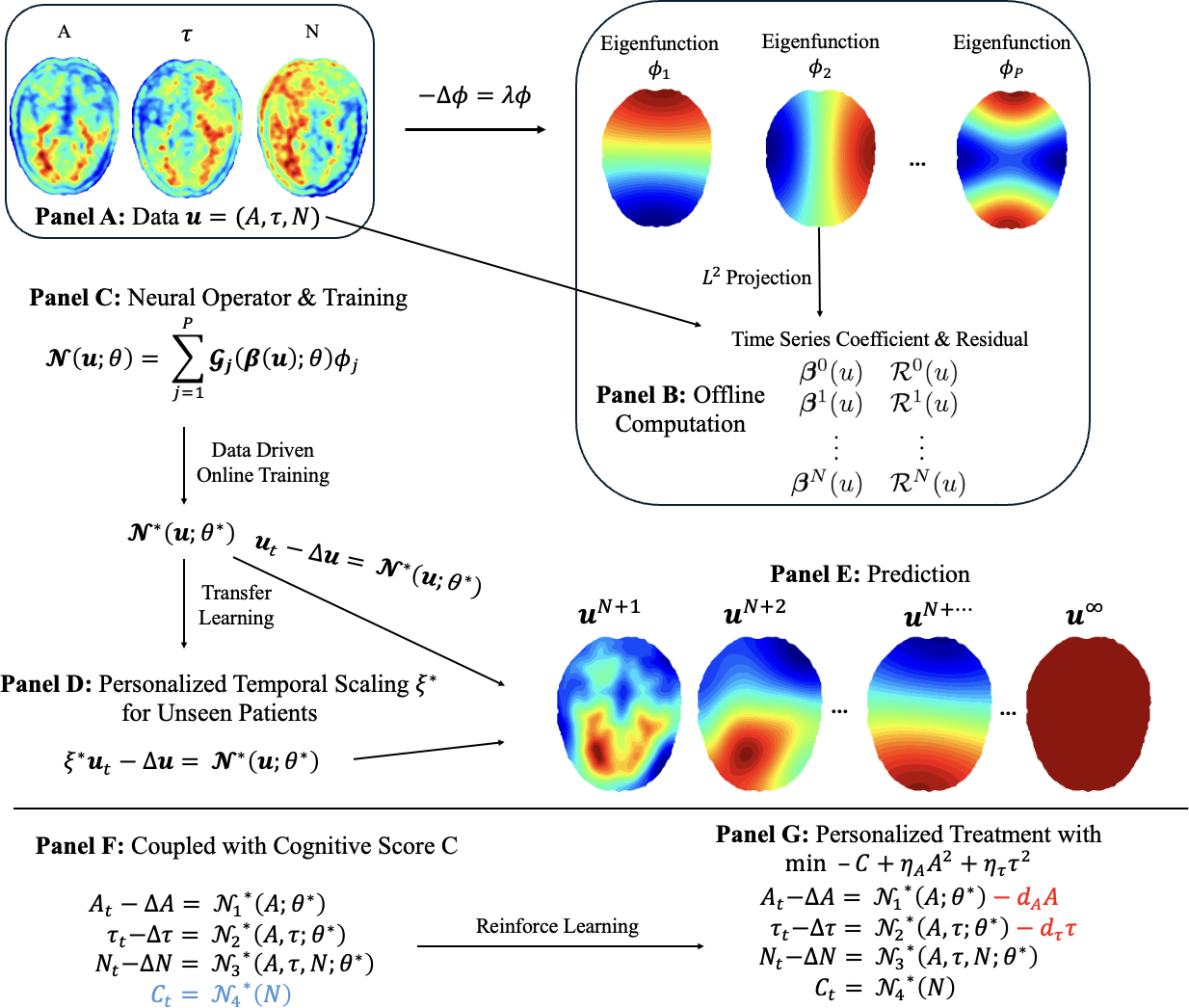}
\caption{Schematic overview of the PDE learning framework using the Laplacian Eigenfunction-Based Neural Operator (LENO). 
\textbf{Panel A:} Clinical imaging data for amyloid-$\beta$ ($A$), tau ($\tau$), and neurodegeneration ($N$) are used as direct inputs. 
\textbf{Panel B:} Laplacian eigenfunctions on the domain $\Omega$ are computed, and imaging data are projected onto these eigenfunctions to obtain time series coefficients $\bm{\beta}^n(u)$ and residuals $\mathcal{R}^n(u)$. 
\textbf{Panel C:} A neural operator $\mathcal{N}(u; \theta)$ is constructed by integrating the eigenfunctions with data-driven online training to approximate the nonlinear terms in the PDE system. 
\textbf{Panel D:}  Transfer learning is applied to unseen patients using a personalized temporal scaling factor $\xi$.
\textbf{Panel E: } The trained operator is then used for forward prediction of biomarker distributions over time.
\textbf{Panel F:}  Neural network learning is extended to incorporate cognitive score $C$, enabling joint modeling of biomarker and cognitive dynamics.
\textbf{Panel G:} Personalized treatment strategies are developed using reinforcement learning to optimize cognitive outcomes.}
    \label{fig:nn}
\end{figure}

Our goal is to directly learn the underlying spatiotemporal dynamics of AD biomarkers from clinical PET imaging data by inferring the governing PDEs that describe the propagation of amyloid-$\beta$ ($A$), tau ($\tau$), and neurodegeneration ($N$) across the brain. 
To achieve this, we employed a Laplacian Eigenfunction-Based Neural Operator (LENO) framework (Fig.~\ref{fig:nn}) that integrates Laplacian eigenfunctions with neural operators to learn the nonlinear progression functions governing biomarker dynamics. The model is trained using imaging data acquired at multiple time points from individual patients.
The LENO model achieved high accuracy in recovering the spatiotemporal dynamics for all three biomarkers. During both training and prediction phases, relative $L^2$ accuracy ($Acc_2$) exceeded 89\% for $A$, $\tau$, and $N$, as summarized in Table~\ref{tab:combined_accuracy}.
To accommodate heterogeneity in disease progression rates across patients, we incorporated a personalized temporal scaling parameter within a transfer learning strategy. This approach preserved strong predictive accuracy for unseen patients (Table~\ref{tab:combined_accuracy}), confirming the framework’s ability to generalize to diverse disease trajectories while learning directly from imaging data.

More specifically, we model the dynamics of $A$-$\tau$-$N$ pathology based on the amyloid cascade hypothesis by formulating a coupled system of PDEs:
\begin{equation}
	\label{eq:atauN}
	\left\{	\begin{aligned}
		A_t - \alpha_A \Delta A & = \mathcal{F}_1(A) && \text{in } \Omega \times (0,T)\\
		\tau_t - \alpha_\tau \Delta \tau & = \mathcal{F}_2(A,\tau) && \text{in } \Omega \times (0,T)\\
		N_t - \alpha_N \Delta N & = \mathcal{F}_3(A,\tau,N) && \text{in } \Omega \times (0,T)
	\end{aligned}\right. \hbox{~with~}    
	\left\{	\begin{aligned}
			A(x,0) &= A_0(x) && \text{in } \Omega\\
		\tau(x,0) &= \tau_0(x) && \text{in } \Omega\\
		N(x,0) &= N_0(x) && \text{in } \Omega.
	\end{aligned}\right.
\end{equation}
Here, $\Delta$ denotes the Laplacian diffusion operator defined on the brain geometry, and $\alpha_A$, $\alpha_\tau$, and $\alpha_N$ are the corresponding diffusion coefficients for $A$, $\tau$, and $N$, respectively. When operating on brain network data, $\Delta$ represents the graph Laplacian constructed from anatomical connectivity between brain regions. 

We adopt a sequential learning strategy based on the LENO framework, guided by the amyloid-$\beta$ (A$\beta$) cascade hypothesis, which posits that the accumulation of A$\beta$ initiates a pathological cascade leading to tau aggregation, neurodegeneration, and ultimately cognitive decline. Reflecting this causal ordering, we first learn the operator $\mathcal{F}_1$ from A$\beta$ ($A$) imaging data, capturing its intrinsic spatiotemporal dynamics. Subsequently, we train $\mathcal{F}_2$ using both $A$ and tau ($\tau$) data to model the downstream influence of A$\beta$ on tau pathology. Finally, we learn $\mathcal{F}_3$ from $A$, $\tau$, and neurodegeneration ($N$) data to characterize the combined effects of upstream pathologies on neurodegenerative progression. During training, the initial few time points are used to estimate both the system’s dynamics and diffusion coefficients, with forward prediction performed at later time points to assess model generalization.

{\bf Evaluation Metrics}
To quantitatively assess model performance, we employ the following accuracy metrics:
\begin{itemize}
    \item \textbf{Relative $L^2$ Accuracy ($Acc_2$):} \quad Measures the normalized $L^2$-norm error between the model prediction $\tilde{u}$ and the ground truth $u$, defined as $    Acc_2 = 1 - \frac{\|u - \tilde{u}\|_{\ell^2}}{\|u\|_{\ell^2}}.$
    \item \textbf{Mean Relative Accuracy ($Acc_1$):} \quad Computes the average relative error across all spatial locations, defined as
 $   Acc_1 = 1 - \frac{1}{|\Omega|} \left\| \frac{u - \tilde{u}}{u} \right\|_{\ell^1},$   where $|\Omega|$ denotes the total number of spatial points.
\end{itemize}

\subsection{Learning on Synthetic Data Simulated from 2D Brain-Shaped Geometries}

\begin{figure}[!htbp]
    \centering

    \begin{subfigure}[t]{0.48\linewidth}
        \centering
        \includegraphics[width=\linewidth]{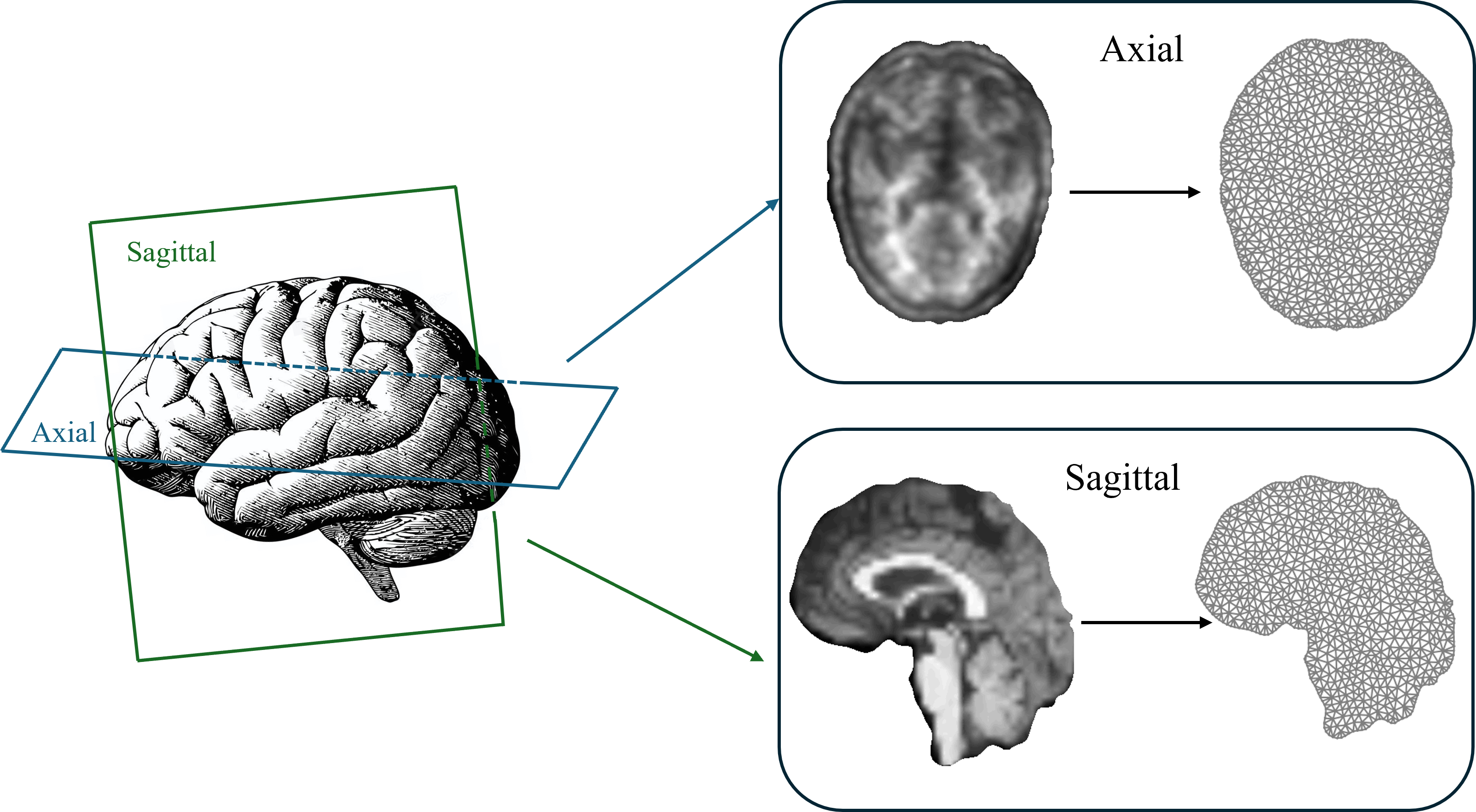}
        \caption*{\textbf{Panel A:} Brain-shaped geometries (Axial and Sagittal) and their corresponding meshes.}
    \end{subfigure}
\hfill
    \raisebox{22mm}{
    \begin{subfigure}[t]{.48\linewidth}
        \centering
        \includegraphics[width=.24\linewidth]{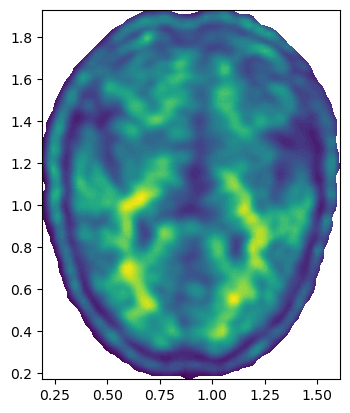}
        \includegraphics[width=.24\linewidth]{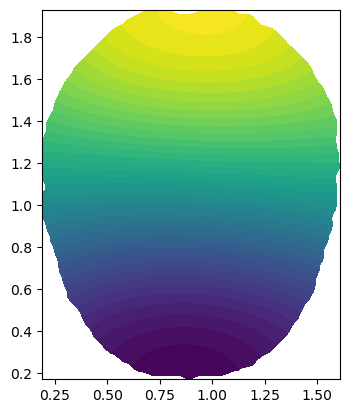}
        \includegraphics[width=.24\linewidth]{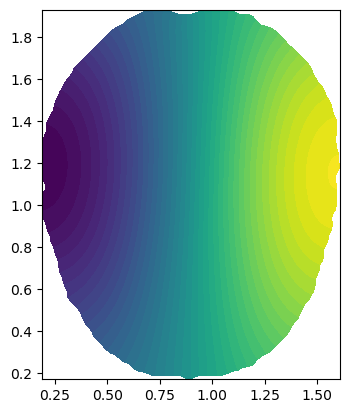}
        \includegraphics[width=.24\linewidth]{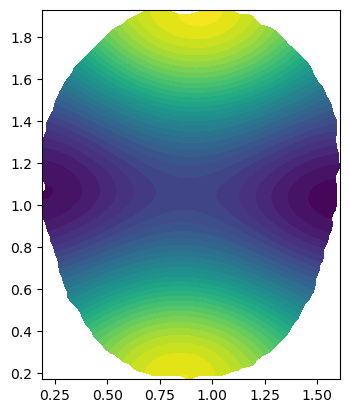}

        \includegraphics[width=.24\linewidth]{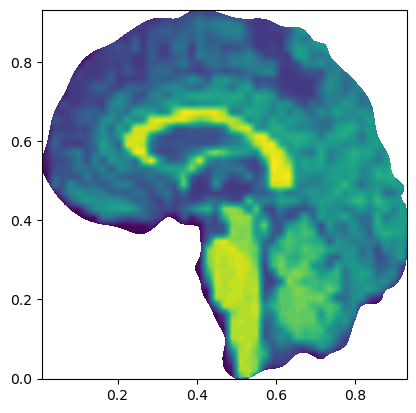}
        \includegraphics[width=.24\linewidth]{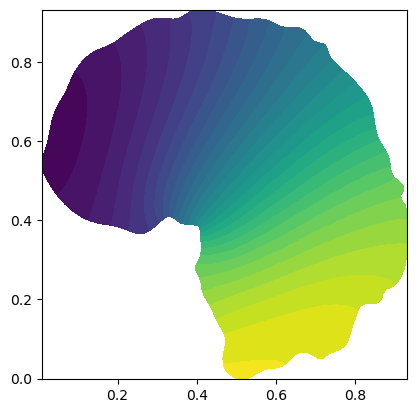}
        \includegraphics[width=.24\linewidth]{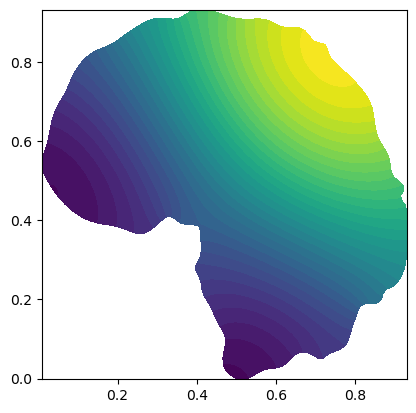}
        \includegraphics[width=.24\linewidth]{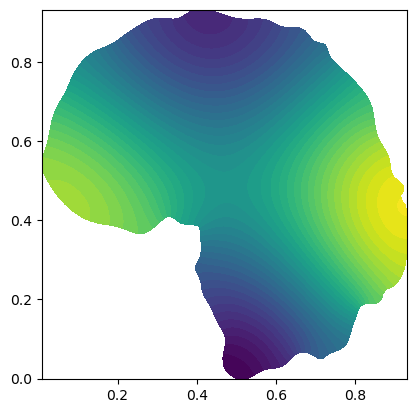}
        \caption*{\textbf{Panel B:} Unstructured brain-shaped domains (Axial: top row, Sagittal: bottom row) and their first three Laplacian eigenfunctions used as spatial basis functions.}
    \end{subfigure}}

    
    \begin{subfigure}[t]{.8\linewidth}
        \centering
        \includegraphics[width=\linewidth]{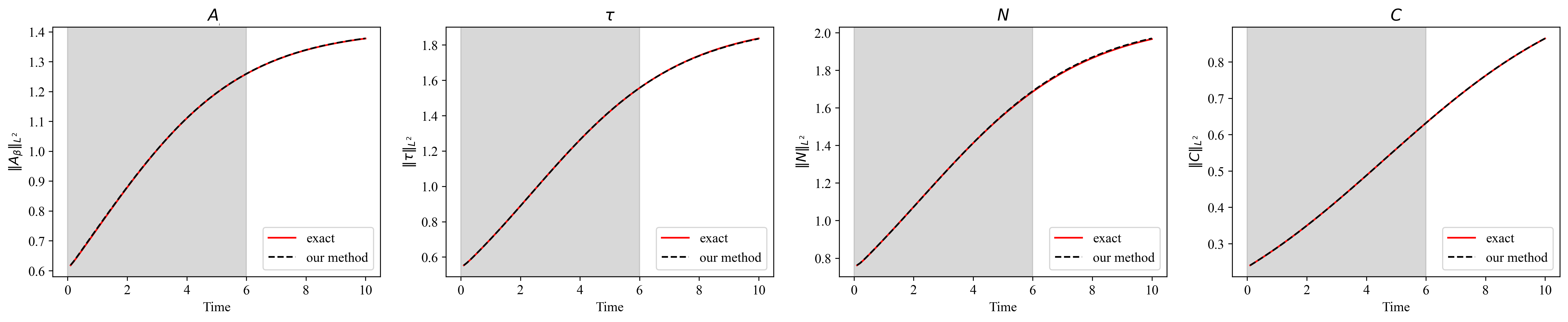}
        \includegraphics[width=\linewidth]{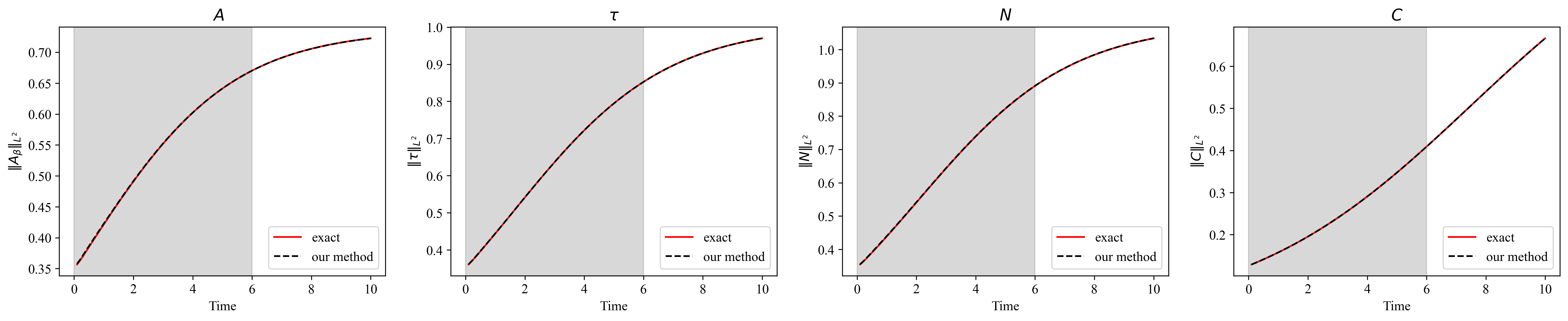}
        \caption*{\textbf{Panel C:} Training and prediction results for Axial (top) and Sagittal (bottom). The shaded region indicates the training interval for one representative initial condition.}
    \end{subfigure}
    \vspace{2mm}

    \begin{subfigure}[t]{.8\linewidth}
        \centering
        \includegraphics[width=.32\linewidth]{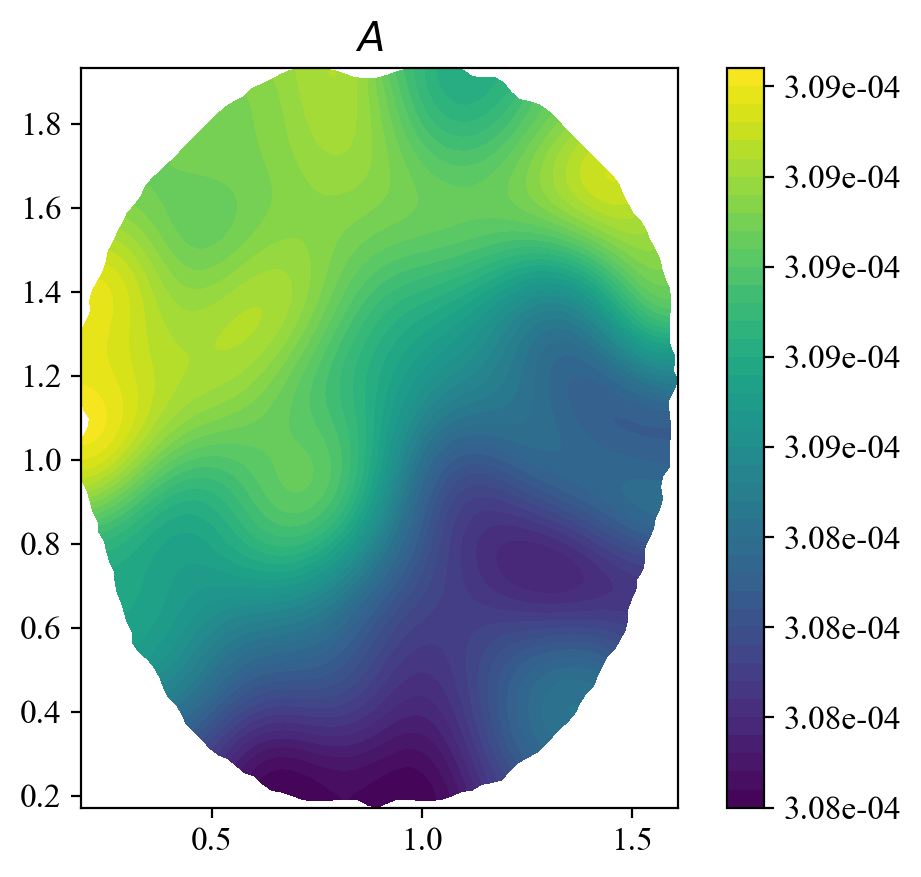}
        \includegraphics[width=.32\linewidth]{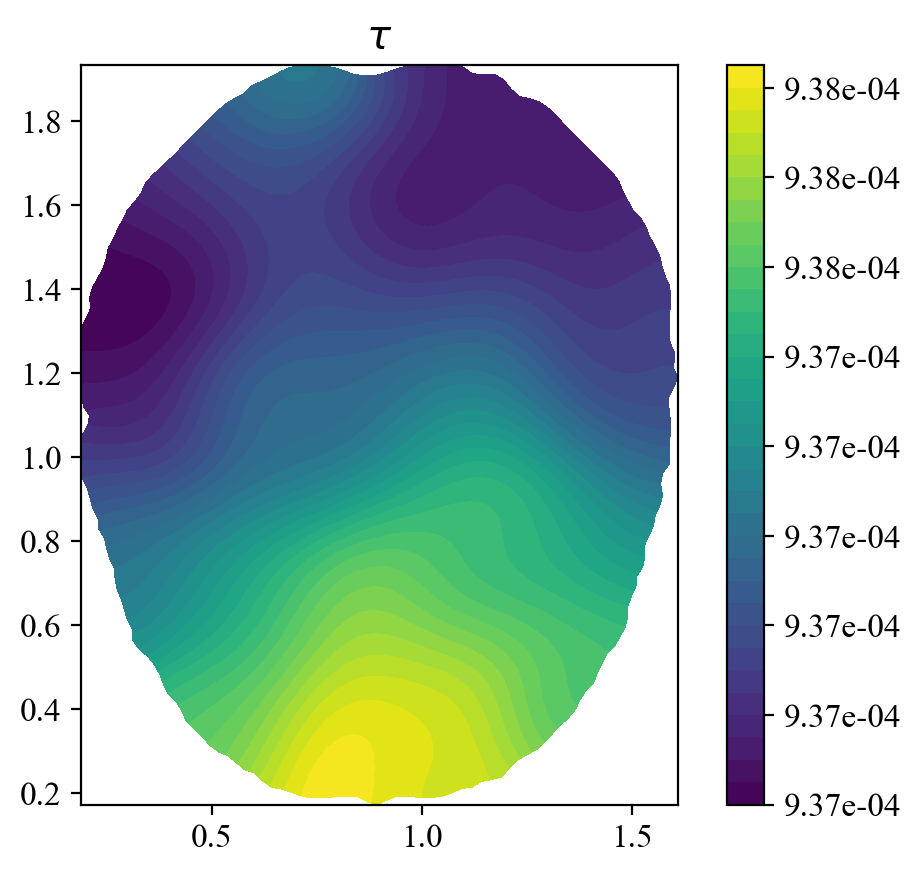}
        \includegraphics[width=.32\linewidth]{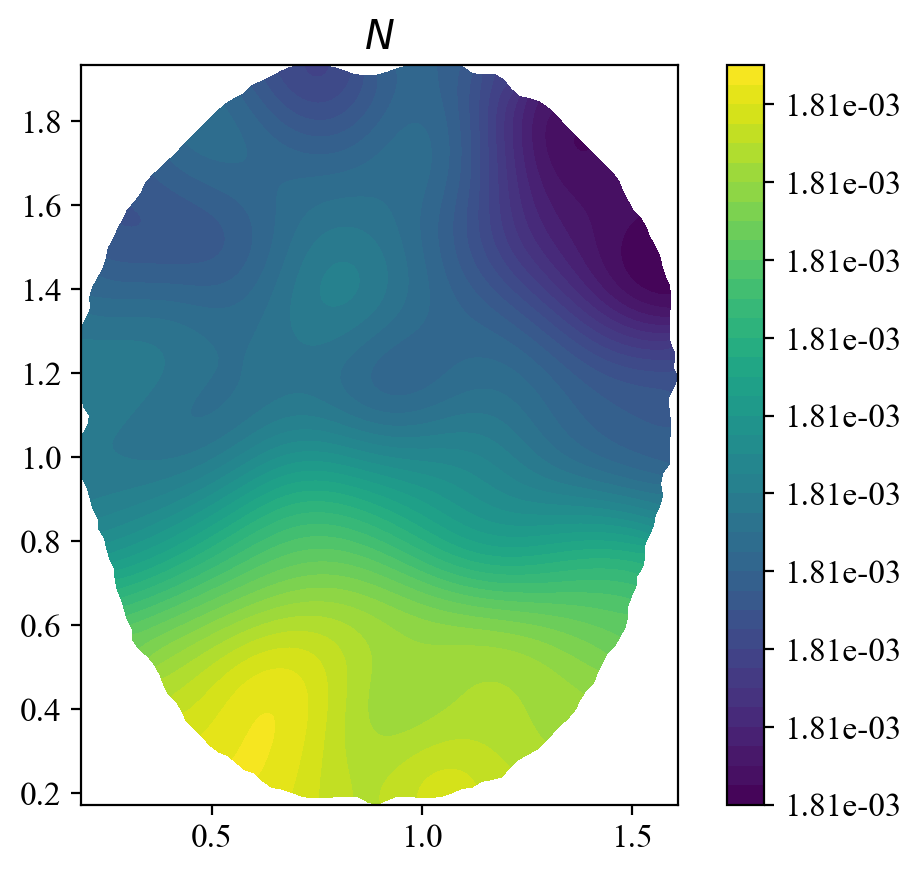}

        \includegraphics[width=.32\linewidth]{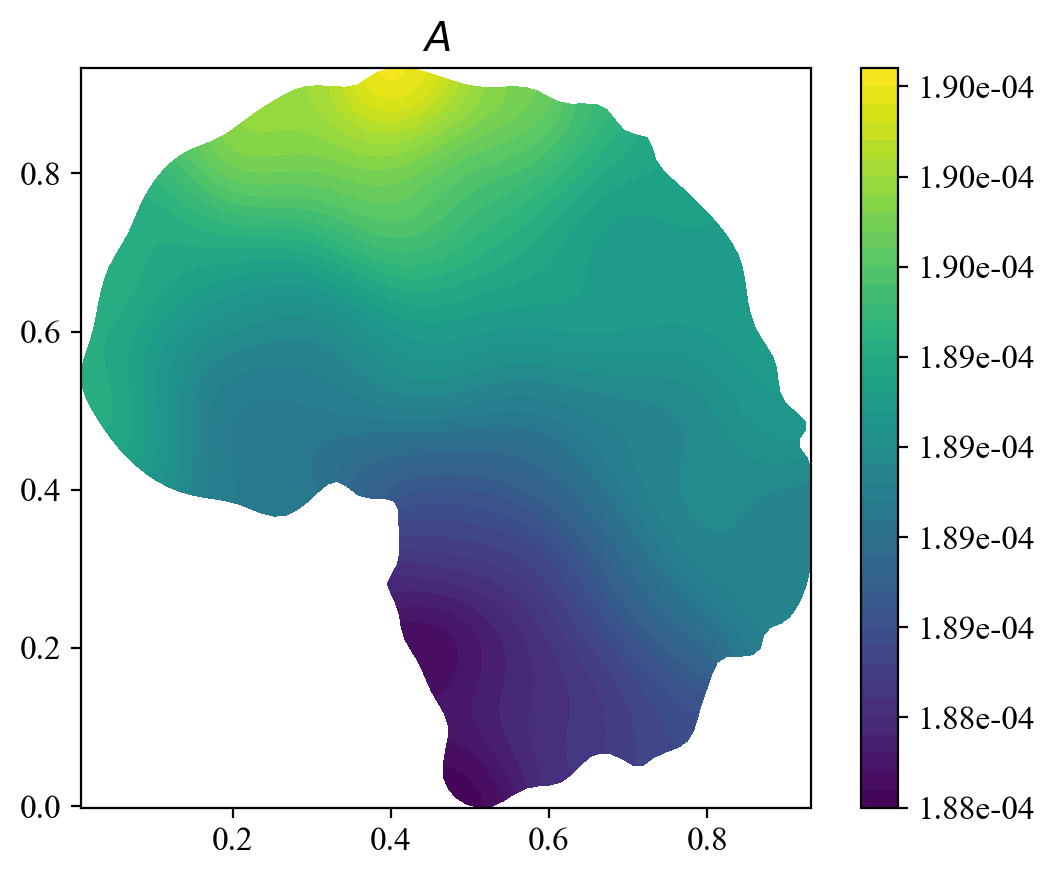}
        \includegraphics[width=.32\linewidth]{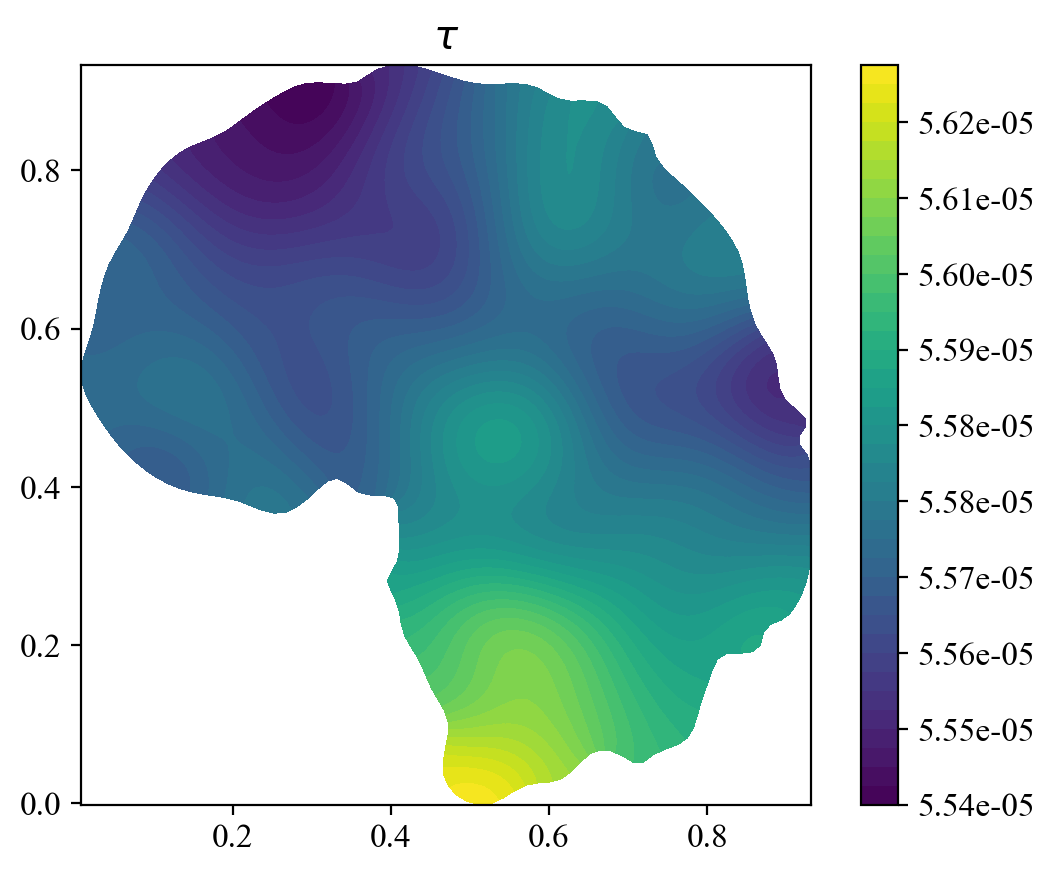}
        \includegraphics[width=.32\linewidth]{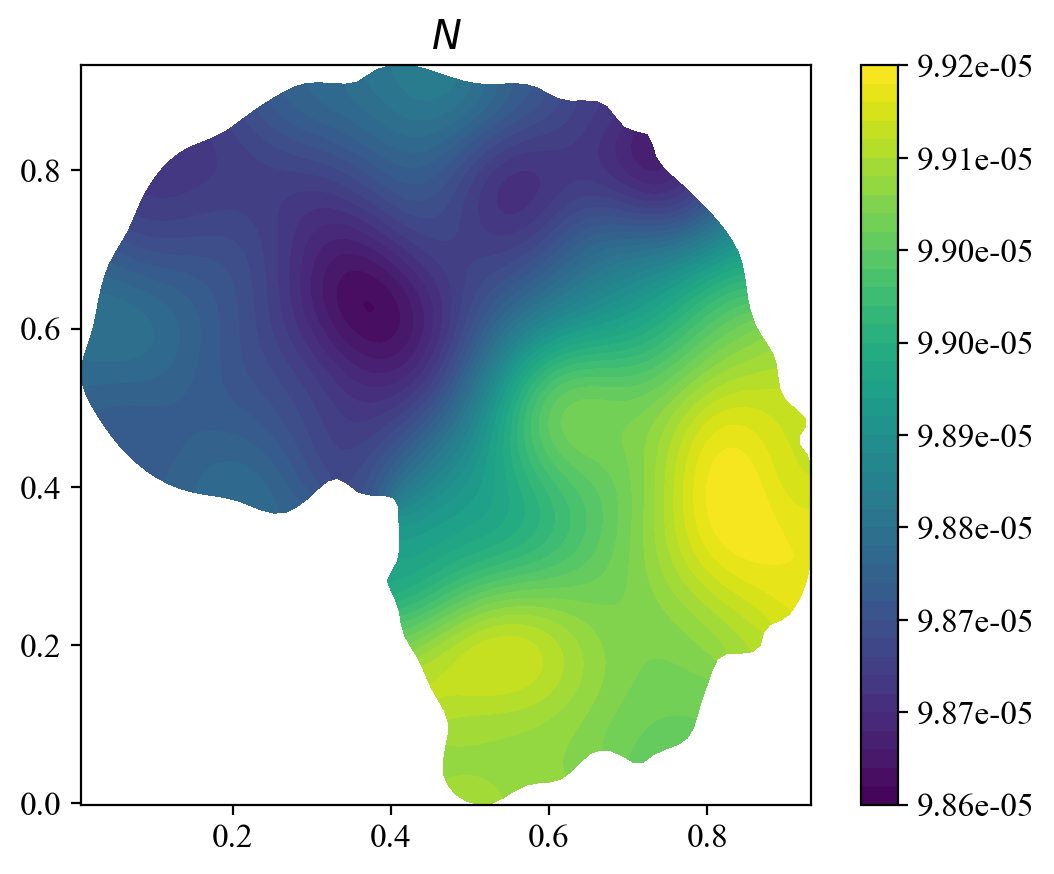}
        \caption*{\textbf{Panel D:} Relative error distributions for $A$, $\tau$, and $N$ on Axial (top row) and Sagittal (bottom row).}
    \end{subfigure}

    \caption{Generalization and performance of the proposed framework on unstructured brain-shaped 2D domains. \textbf{Panel A:} Geometries and meshes. \textbf{Panel B:} Eigenfunctions of the Laplacian used as neural operator basis. \textbf{Panel C:} Training and prediction of biomarkers. \textbf{Panel D:} Relative error maps for each variable.}
    \label{fig:brain_all}
\end{figure}

To assess the flexibility and spatial generalization capability of our framework, we generated synthetic data by simulating the coupled A$\beta$-$\tau$-$N$ pathology dynamics on two unstructured, brain-shaped 2D geometries (referred to as slip 1 and slip 2) in Figure \ref{fig:brain_all}. 


These geometries were designed to approximate realistic cortical boundaries while preserving heterogeneous spatial connectivity patterns.

The synthetic biomarker distributions were obtained by numerically solving a system of reaction-diffusion PDEs (Equation~\ref{eq_sys}) incorporating biologically inspired nonlinear interaction terms and no-flux boundary conditions. 
 Reaction-diffusion PDEs have been widely applied to model neurodegenerative disease progression, as they capture both the local reaction dynamics—such as amyloid accumulation, tau propagation, and neuronal degeneration—and the spatial diffusion of pathology across connected brain regions. 
Specifically, the system of PDEs has the following form \cite{petrella2019computational}:
\begin{equation}
	\left\{
	\begin{aligned}
		\dfrac{dA}{dt} - \alpha_A \Delta A &= \lambda_A A (K_A - A), \\
		\dfrac{d\tau}{dt} - \alpha_\tau \Delta \tau &= \lambda_{\tau A} A + \lambda_\tau \tau (K_\tau - \tau), \\
		\dfrac{dN}{dt} - \alpha_N \Delta N &= \lambda_{N\tau} \tau + \lambda_N N (K_N - N), \\
		\dfrac{dC}{dt} &= \lambda_{CN} \int N + \lambda_C C (K_C - C),
	\end{aligned}
	\right.
	\label{eq_sys}
\end{equation}
subject to no-flux boundary conditions. This modeling approach is consistent with the amyloid cascade hypothesis and prion-like trans-synaptic spreading mechanisms implicated in AD pathology \cite{raj2012network, moravveji2024scoping}.

The initial conditions for each simulation were derived from PET imaging data of 20 patients to provide physiologically plausible baseline biomarker distributions.

To enable PDE learning on these irregular spatial domains, Laplacian eigenfunctions computed for each geometry were employed as spatial basis functions within the LENO framework. This strategy allows the model to approximate unknown nonlinear terms and differential operators directly from the simulated spatiotemporal data. Figure~\ref{fig:brain_all} displays the simulation domains and the first three principal Laplacian eigenfunctions, which serve as the basis functions for the neural operator during training.


\paragraph{Quantitative Results}

{\bf Training errors:} We evaluated the LENO framework on the synthetic A$\beta$-$\tau$-$N$ system described in Equation~\ref{eq_sys}. Table~\ref{tab:brain_combined} reports quantitative performance metrics on both simulation domains (Axial and Sagittal). Specifically, we computed the relative $L^2$ error ($E_{L^2}$), the PDE residual error ($E_{Res}$), and the nonlinear recovery error ($E_{Nonlinear}$).

\begin{table}[!htbp]
\centering
\begin{tabular}{c|cccc|cccc}
\hline
\multirow{2}{*}{Variable} & \multicolumn{4}{c|}{\textbf{Axial}} & \multicolumn{4}{c}{\textbf{Sagittal}} \\
& $E_{L^2}$ & $E_{Res}$ & $E_{Nonlinear}$ & $E_{L^2}^{\text{pred}}$ & $E_{L^2}$ & $E_{Res}$ & $E_{Nonlinear}$ & $E_{L^2}^{\text{pred}}$ \\
\hline
$A$    & 2.08e-04 & 3.73e-03 & 7.65e-02 & 3.09e-04 & 1.36e-04 & 4.99e-03 & 9.19e-02 & 1.89e-04 \\
$\tau$ & 1.89e-04 & 2.96e-03 & 6.00e-02 & 9.37e-04 & 8.75e-05 & 2.75e-03 & 9.25e-02 & 5.57e-05 \\
$N$    & 3.74e-04 & 5.13e-03 & 6.46e-02 & 1.81e-03 & 1.35e-04 & 2.66e-03 & 8.66e-02 & 9.89e-05 \\
$C$    & 1.03e-04 & 1.09e-03 & 1.83e-02 & 2.86e-04 & 5.97e-05 & 7.40e-04 & 1.21e-02 & 2.55e-06 \\
\hline
\end{tabular}
\caption{Error metrics for simulated variables on brain-shaped domains. Relative $L^2$ error ($E_{L^2}$), PDE residual error ($E_{Res}$), nonlinear recovery error ($E_{Nonlinear}$), and prediction error at $t = 10$ ($E_{L^2}^{\text{pred}}$) are shown for each variable under \textbf{Axial} and \textbf{Sagittal}.}\label{tab:brain_combined}
\end{table}


{\bf Prediction errors:} We further report the mean relative $L^2$ prediction error at time $t=10$ for each simulated variable in Table~\ref{tab:brain_combined}. The model achieves consistently low errors across both domains, demonstrating robustness to domain geometry.


{\bf Temporal Prediction and Spatial Error Distribution:} Figure~\ref{fig:brain_all} illustrates the temporal trajectories of $A$, $\tau$, $N$, and $C$ for both simulation domains. The shaded region indicates the training window, while predictions extend beyond this interval, validating the model’s forecasting ability.
We also visualize the relative error distributions for $A$, $\tau$, and $N$ in Figure~\ref{fig:brain_all}, confirming spatially consistent error profiles across both domains.



\subsection{Learning on SUVR Data from 3D Brain Networks}

To evaluate the applicability of our framework on real neuroimaging data, we applied the LENO framework to longitudinal SUVR measurements of amyloid-$\beta$ ($A$), tau ($\tau$), and neurodegeneration ($N$) collected from \textbf{100 patients}. Each patient had between 3–5 imaging time points spanning several years, capturing heterogeneous disease progression patterns.

The SUVR maps were spatially normalized and parcellated according to the Desikan-Killiany atlas, resulting in \textbf{$N=68$ cortical regions} per patient. To incorporate individual-specific anatomical constraints, we constructed personalized brain networks using diffusion MRI tractography data for each patient. The anatomical connectivity was represented by an adjacency matrix $W$, where $W_{ij}$ denotes the fiber connection strength between brain regions $i$ and $j$. From this, the graph Laplacian was computed as
$    L = D - W,$ with $D$ being the diagonal degree matrix defined by $D_{ii} = \sum_j W_{ij}$.

We then computed the first \textbf{$K=48$ Laplacian eigenfunctions} for each patient-specific brain network. These eigenfunctions act as spatial basis functions in the LENO model, effectively encoding the underlying topological constraints of the brain's network structure and supporting the learning of spatiotemporal biomarker dynamics. The number of eigenfunctions was selected based on a trade-off between accuracy and computational cost, and this choice yielded stable training and reliable predictions in both synthetic and patient data experiments.

We trained our model using the available longitudinal SUVR data for each patient, reserving the later time points for prediction. To quantify model performance, we computed relative accuracy metrics for both training and prediction phases.

\paragraph{Training and Prediction Accuracy}

Table~\ref{tab:combined_accuracy} summarizes the relative training and prediction accuracies for each biomarker. The model consistently achieves high accuracy across both metrics, validating its effectiveness on irregular brain-shaped geometries.

\begin{table}[!htbp]
	\centering
	\begin{tabular}{c|cc||cc}
    \hline
    \multicolumn{5}{c}{Training and prediction accuracy for $A$, $\tau$, and $N$ on training patients}\\
		\hline
		Variable & Training $Acc_{2}$ (\%) & Training $Acc_{1}$ (\%) & Prediction $Acc_{2}$ (\%) & Prediction $Acc_{1}$ (\%) \\
		\hline
		$A$    & 92.56 & 93.75 & 89.64 & 90.92 \\
		$\tau$ & 95.20 & 96.11 & 93.69 & 94.76 \\
		$N$    & 94.58 & 95.98 & 93.88 & 95.33 \\
		\hline
	\end{tabular}

    \begin{tabular}{c|cc||cc}
		\hline
        \multicolumn{5}{c}{Training and prediction accuracy for $A$, $\tau$, and $N$ on new patients via transfer learning}\\
        \hline
		Variable & Training $Acc_{2}$ (\%) & Training $Acc_{1}$ (\%) & Prediction $Acc_{2}$ (\%) & Prediction $Acc_{1}$ (\%) \\
		\hline
		$A$    & 91.63 & 92.99 & 89.63 & 91.15 \\
		$\tau$ & 94.50 & 95.64 & 92.55 & 94.05 \\
		$N$    & 94.39 & 95.52 & 93.98 & 95.16 \\
		\hline
	\end{tabular}
 \setlength{\tabcolsep}{10mm}
    \begin{tabular}{c|c||c}
   
    \hline
    \multicolumn{3}{c}{Training and prediction accuracy for cognitive score \( C \).}\\
		\hline
		Variable & Training  Accuracy $Acc_2$ (\%) & Prediction Accuracy $Acc_2$ (\%) \\
		\hline
		$C$ & 98.59 & 94.07 \\
		\hline
	\end{tabular}
    	\caption{Training and prediction accuracy for $A$, $\tau$, $N$, and cognitive score $C$. The first two tables report the training and prediction accuracy for $A$, $\tau$, and $N$ in (i) training patients and (ii) patients via transfer learning, using accuracy metrics $Acc_2$ and $Acc_1$. The third table presents the training and prediction accuracy for the cognitive score $C$ using $Acc_2$}	\label{tab:combined_accuracy}
\end{table}

\paragraph{Transfer Learning Across Patients}

To accommodate inter-patient heterogeneity in disease progression rates, we applied transfer learning by fixing the trained network weights and introducing a patient-specific disease progression score $s = \gamma t + \beta$. The scaling parameter $\gamma$ was optimized for each new patient, while the model structure remained unchanged. Table~\ref{tab:combined_accuracy} reports the resulting training and prediction accuracies, demonstrating effective generalization to new patients through simple temporal rescaling.

\begin{figure}[!htbp]
    \centering
    \begin{subfigure}[b]{.45\textwidth}
        \centering
          \includegraphics[width=.7\textwidth]{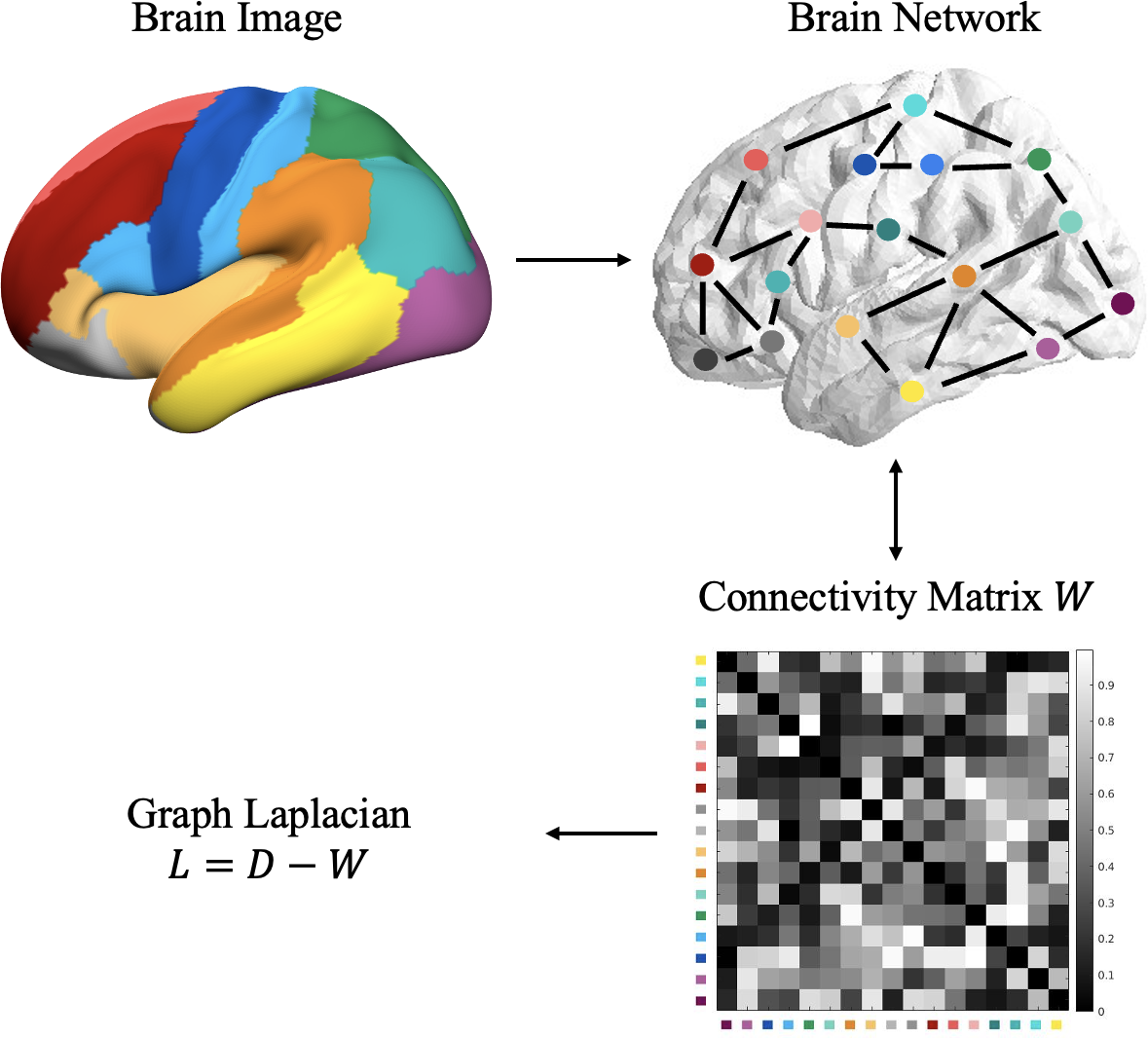}
        \caption*{\textbf{Panel A:} Image preprocessing pipeline for constructing the graph Laplacian matrix $L$.}
    \end{subfigure}
    \hfill
    \raisebox{4mm}{
    \begin{subfigure}[b]{.5\textwidth}
        \centering
        \includegraphics[width=0.4\textwidth]{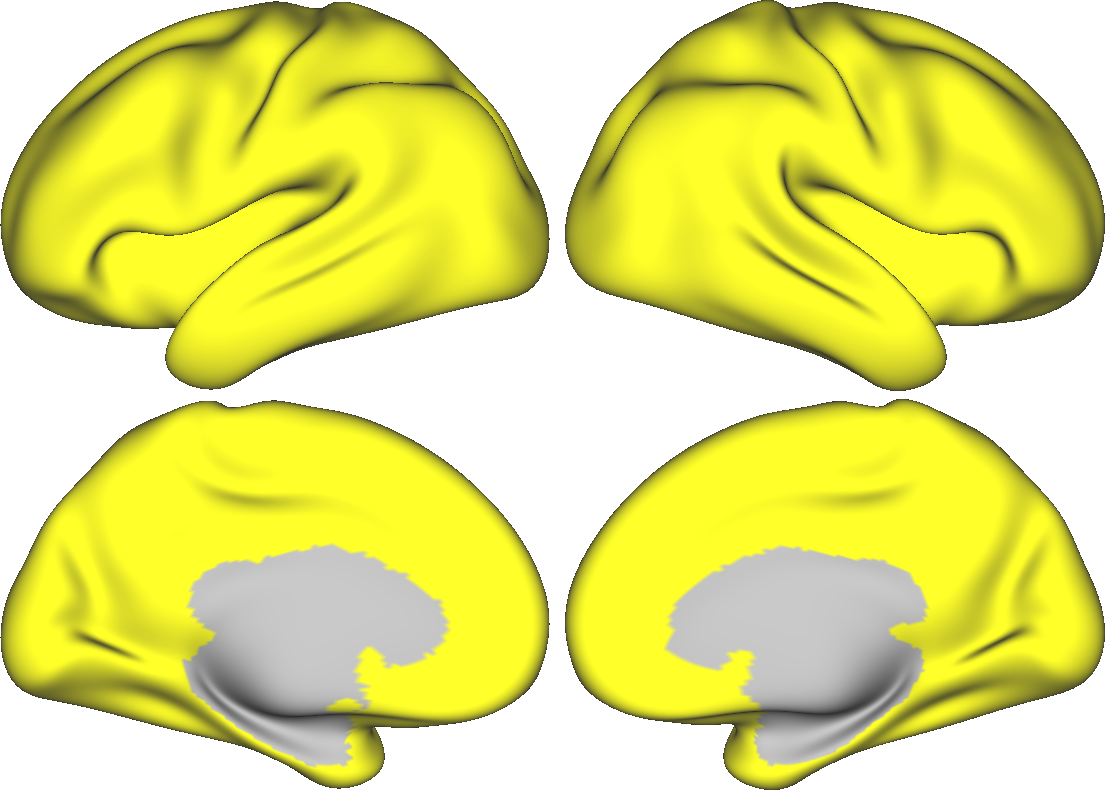}
         \includegraphics[width=0.4\textwidth]{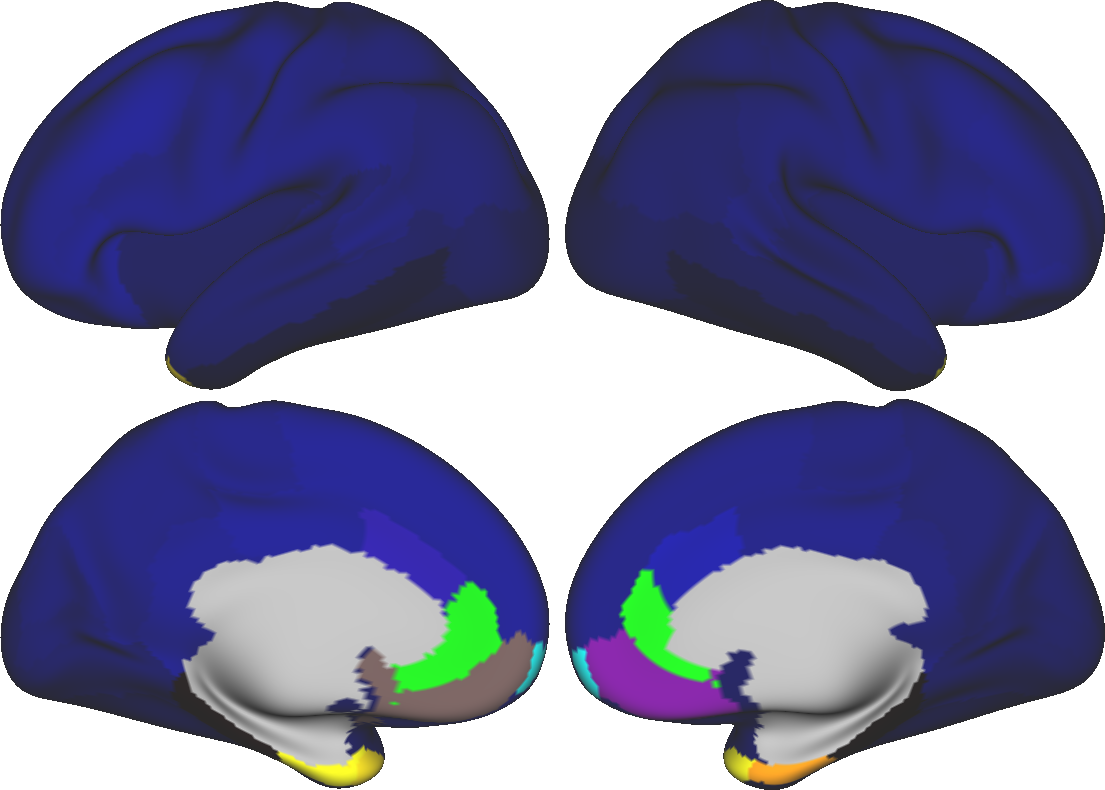}
         \\
          \includegraphics[width=0.4\textwidth]{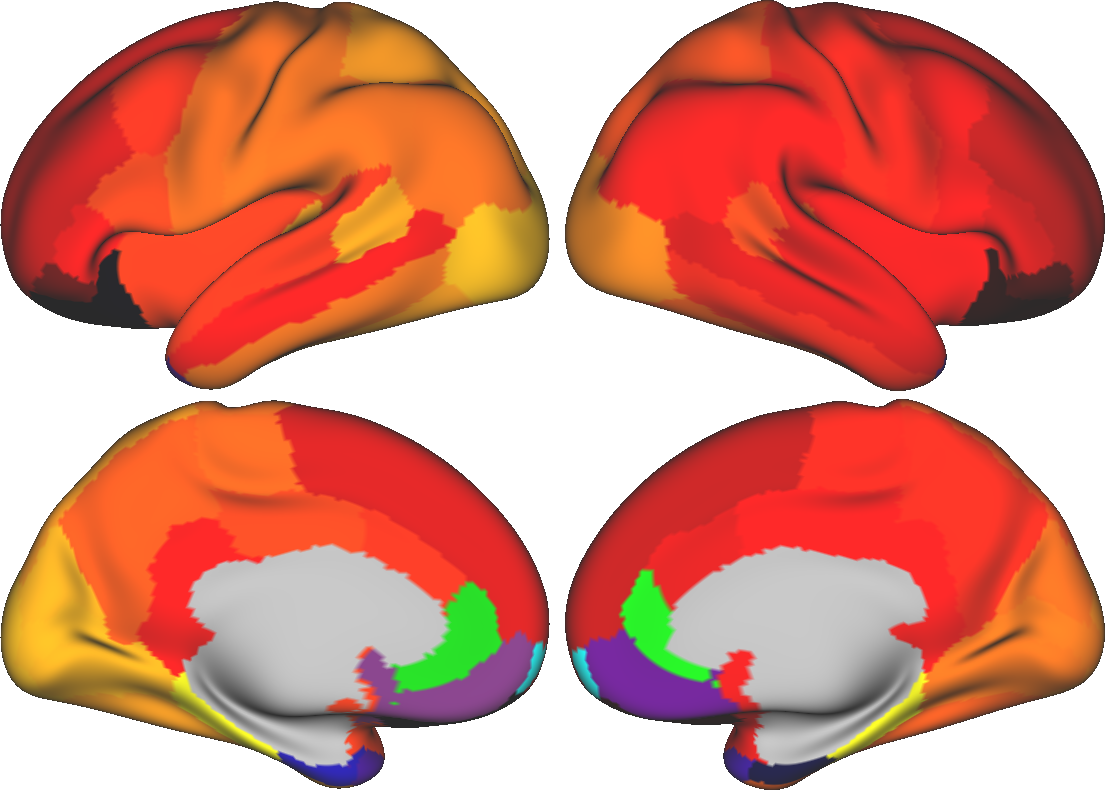}
          \includegraphics[width=0.4\textwidth]{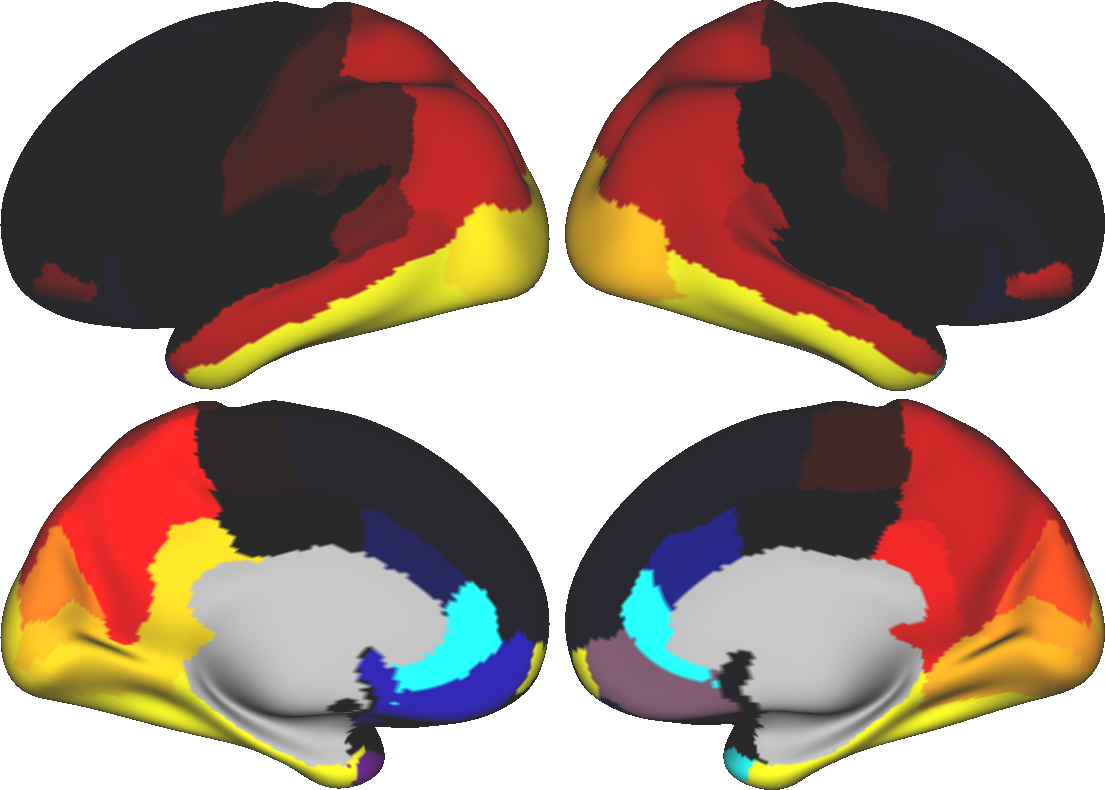}
        \caption*{\textbf{Panel B:} Graph Laplacian eigenfunctions in the 3D brain.}
    \end{subfigure}}\\

    \vspace{5mm}

    \begin{subfigure}[!htbp]{1\textwidth}
        \centering
        \begin{tabular}{ccc||cc||cc}
            \hline
            Biomarker 
            &\multicolumn{2}{|c||}{\textbf{$A$}} &\multicolumn{2}{c||}{\textbf{$\tau$}}
            &\multicolumn{2}{c}{\textbf{$N$}}\\ 
            \hline
            \multicolumn{1}{c|}{Age} & 67 & 69 & 70 & 71 & 71 & 72\\
            \hline
            \multicolumn{1}{c|}{LENO}
            &\includegraphics[width=0.12\textwidth,valign=c]{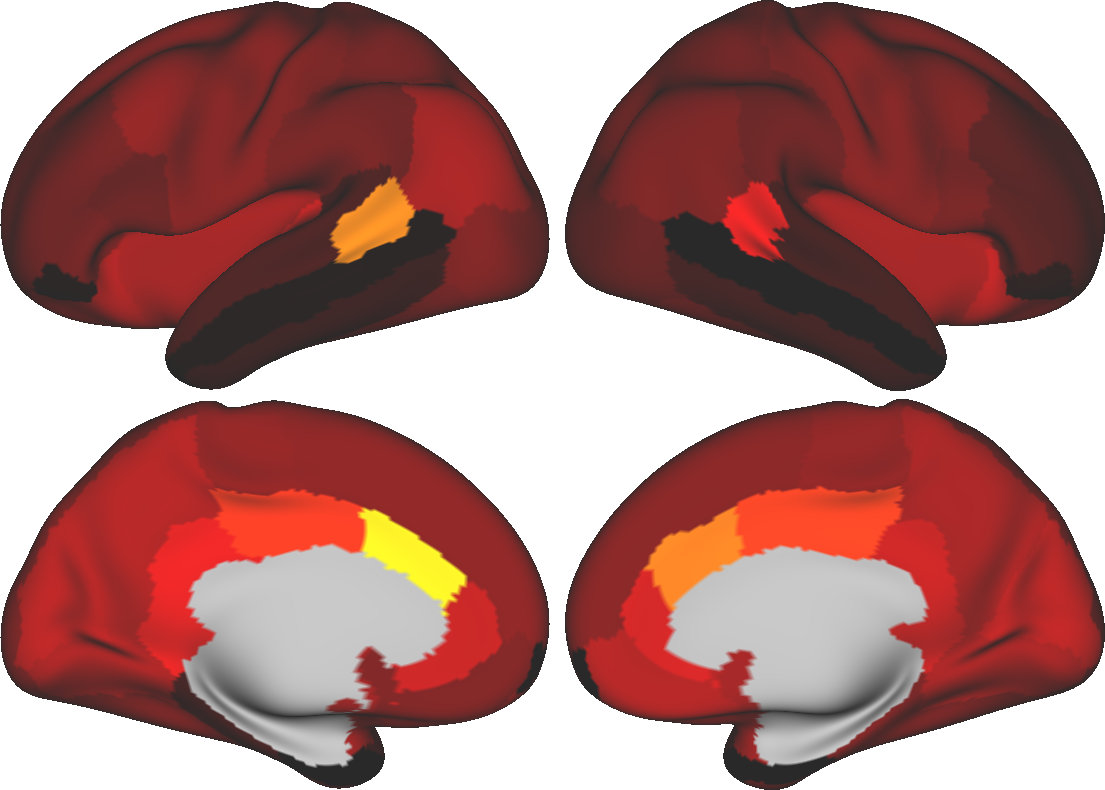} &
            \includegraphics[width=0.12\textwidth,valign=c]{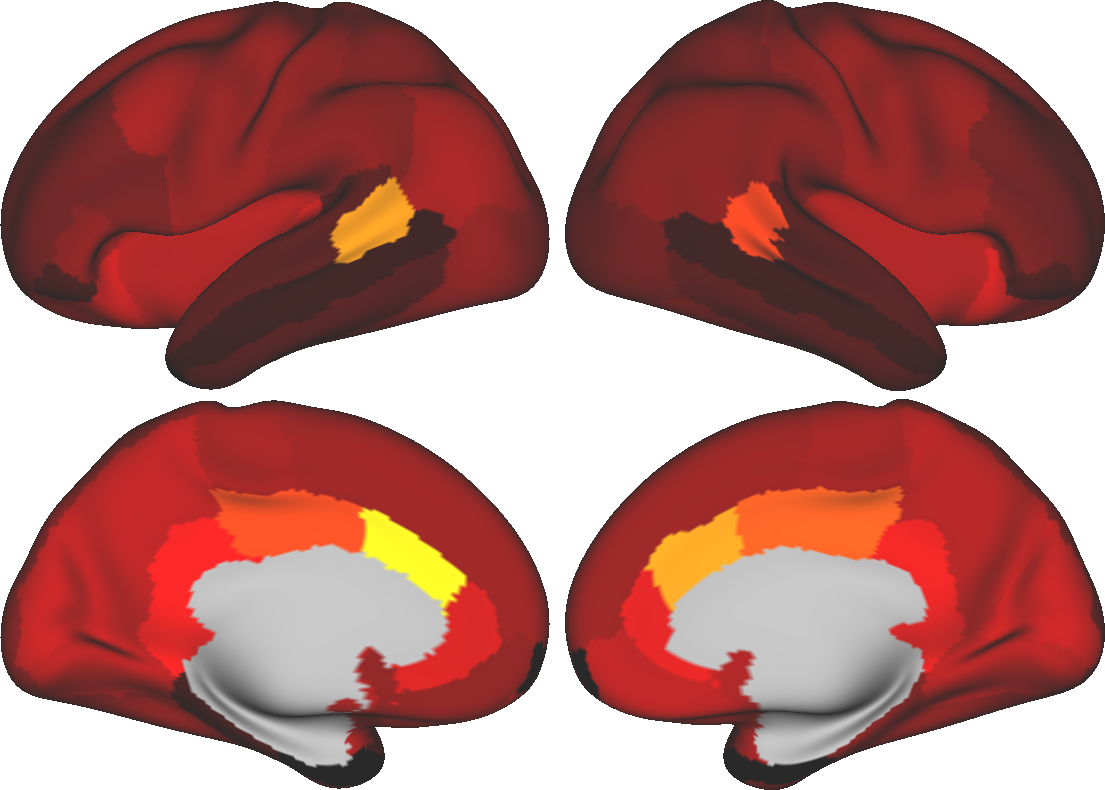} &
            \includegraphics[width=0.12\textwidth,valign=c]{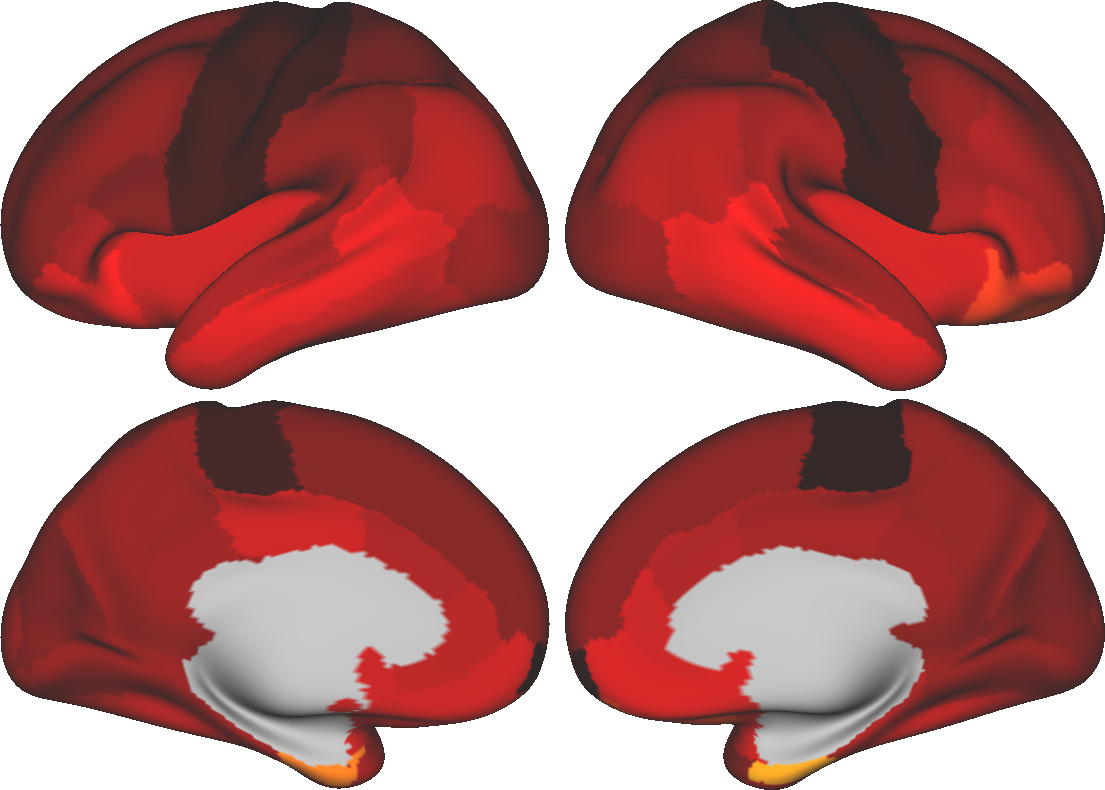} &
            \includegraphics[width=0.12\textwidth,valign=c]{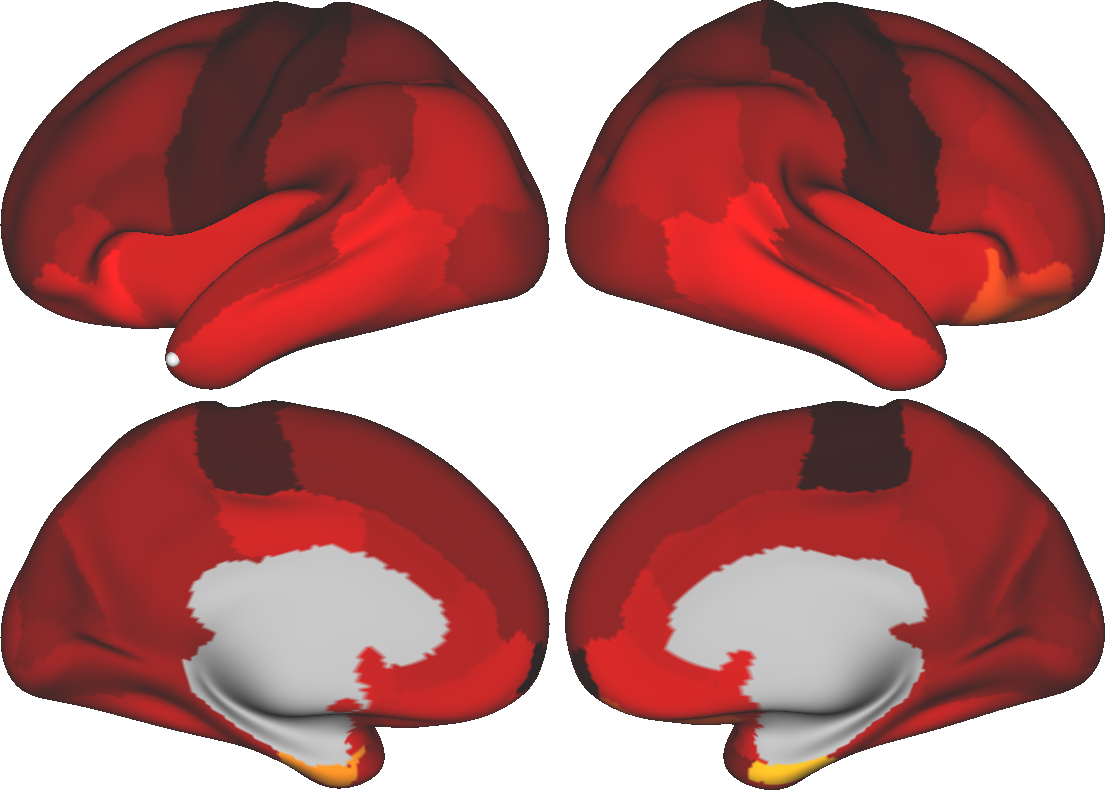} &
            \includegraphics[width=0.12\textwidth,valign=c]{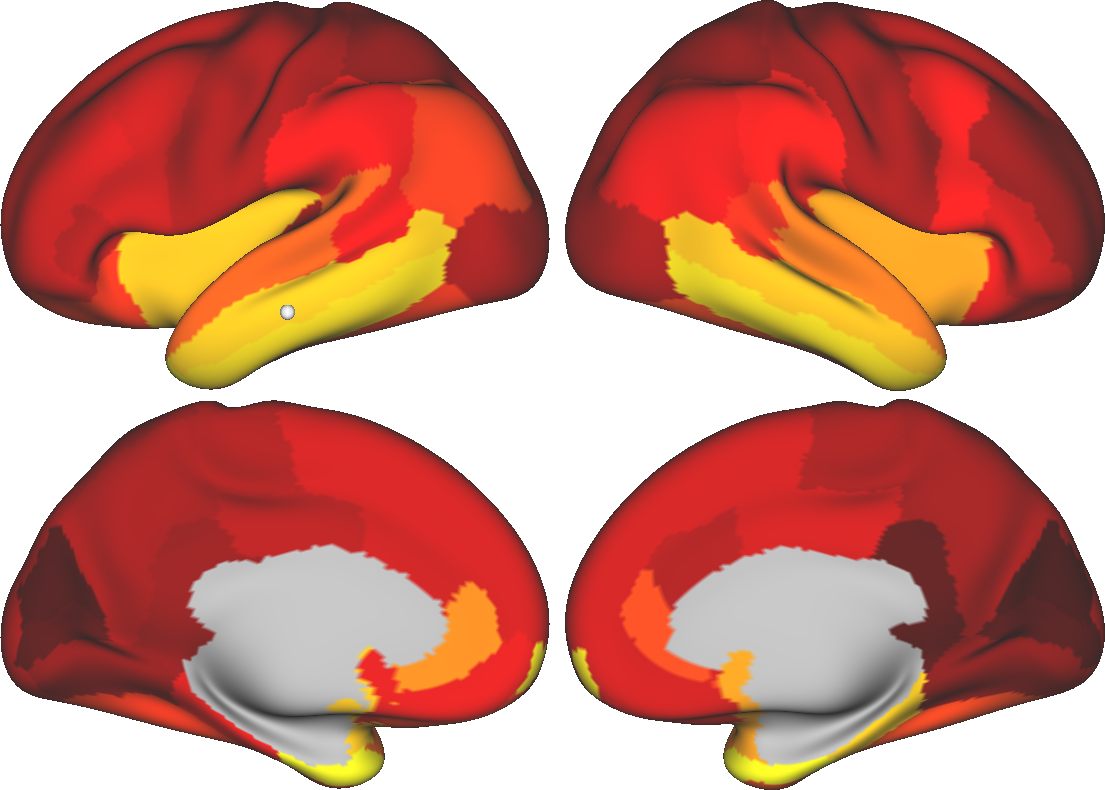} &
            \includegraphics[width=0.12\textwidth,valign=c]{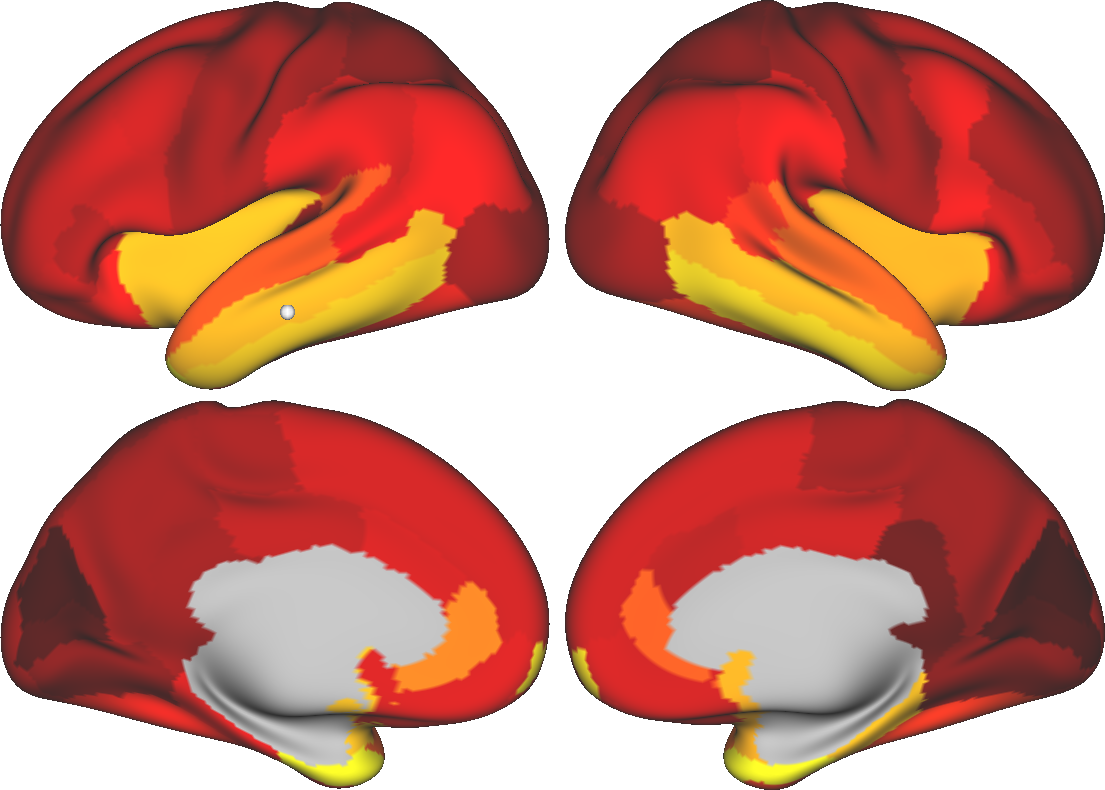} \\
            \hline
            \multicolumn{1}{c|}{Truth} &\includegraphics[width=0.12\textwidth,valign=c]{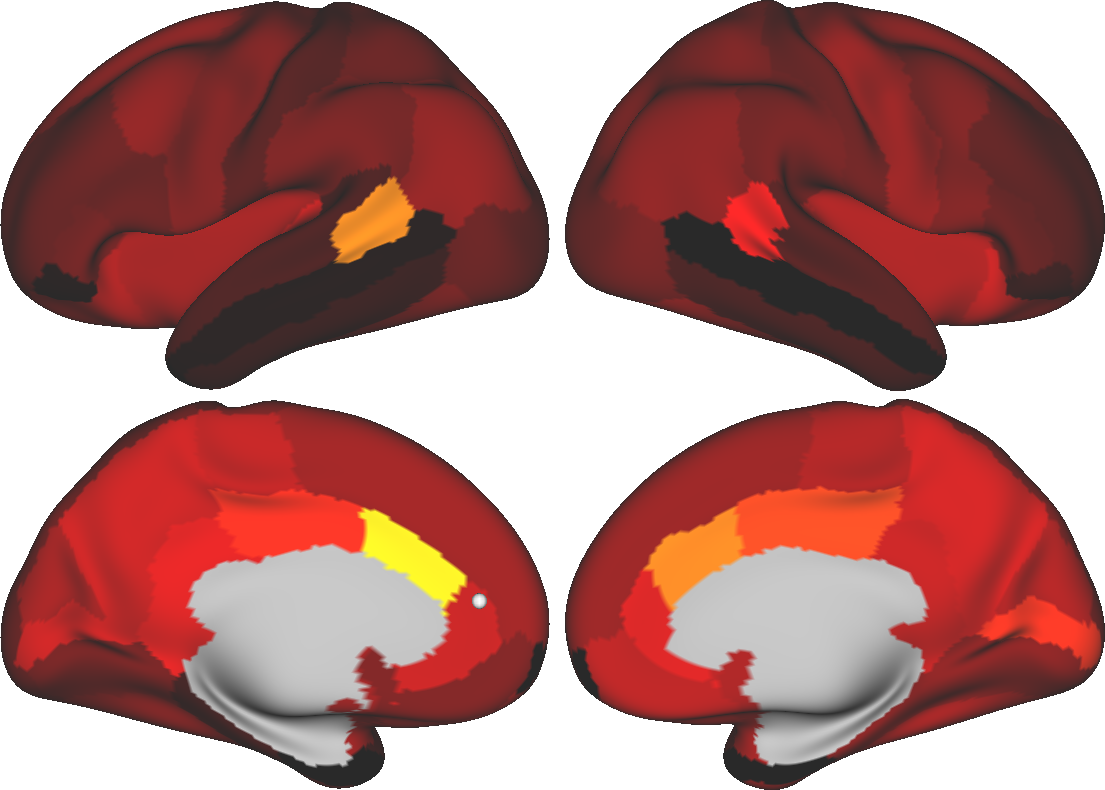} &
            \includegraphics[width=0.12\textwidth,valign=c]{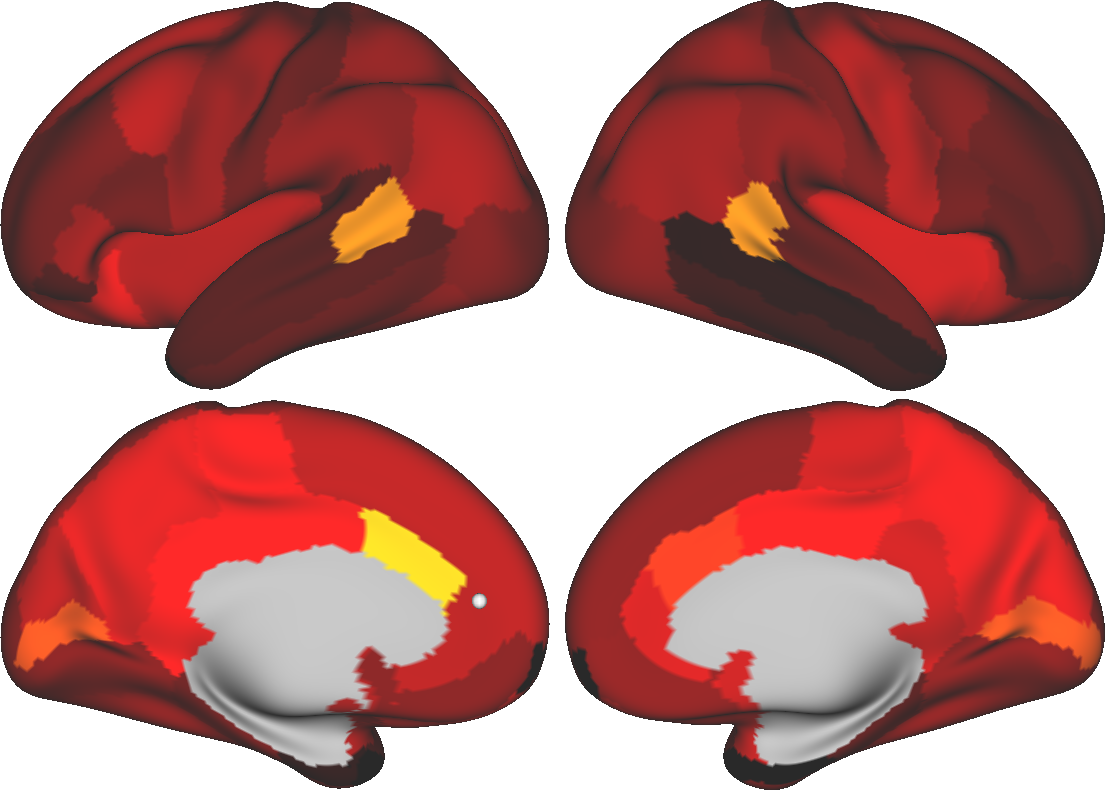} &
            \includegraphics[width=0.12\textwidth,valign=c]{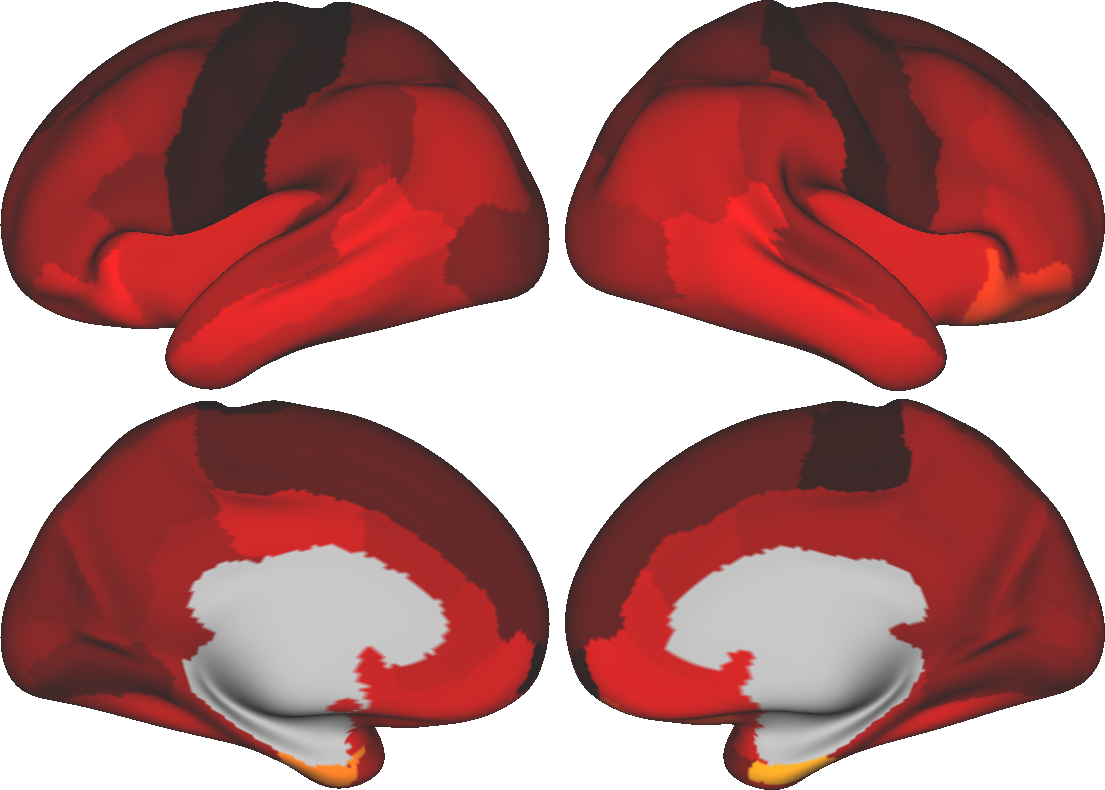} &
            \includegraphics[width=0.12\textwidth,valign=c]{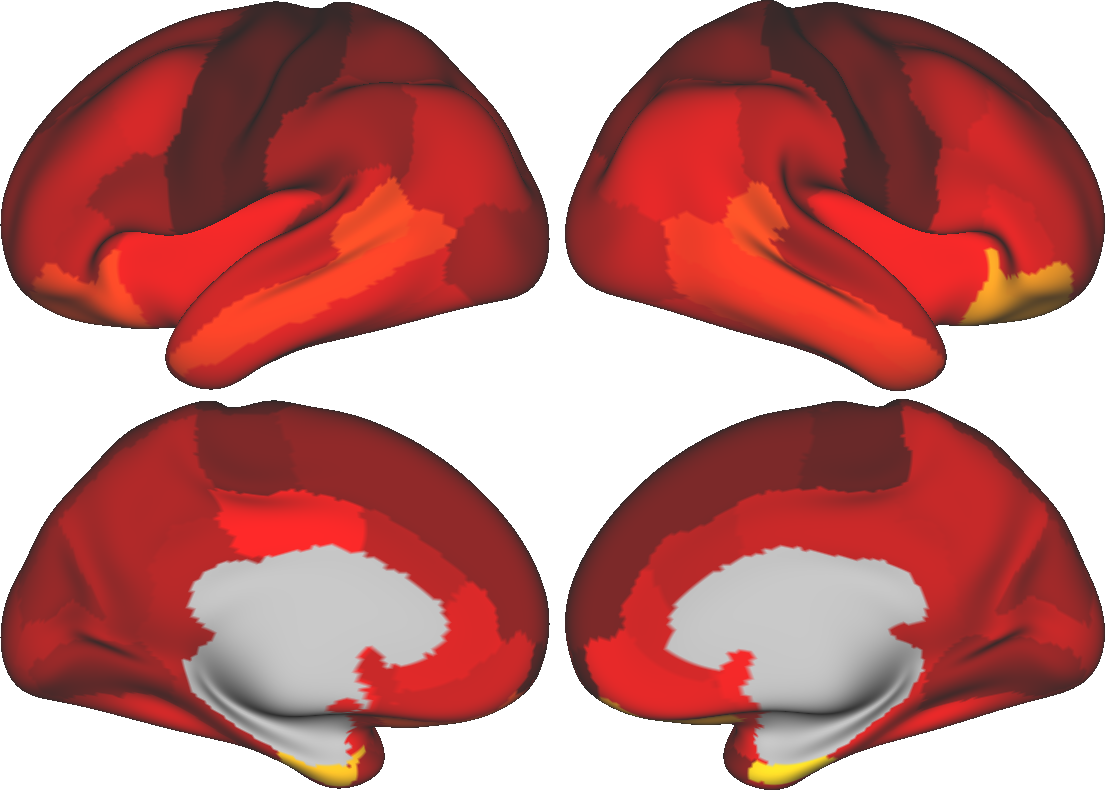} &
            \includegraphics[width=0.12\textwidth,valign=c]{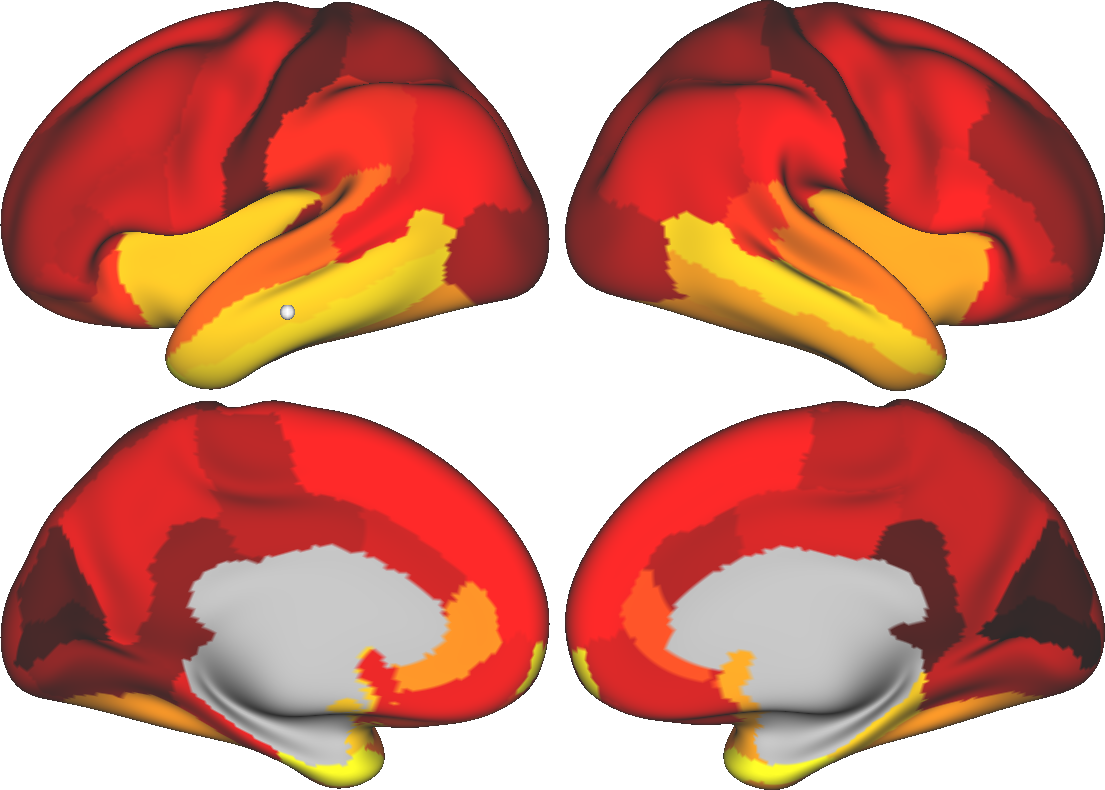} &
            \includegraphics[width=0.12\textwidth,valign=c]{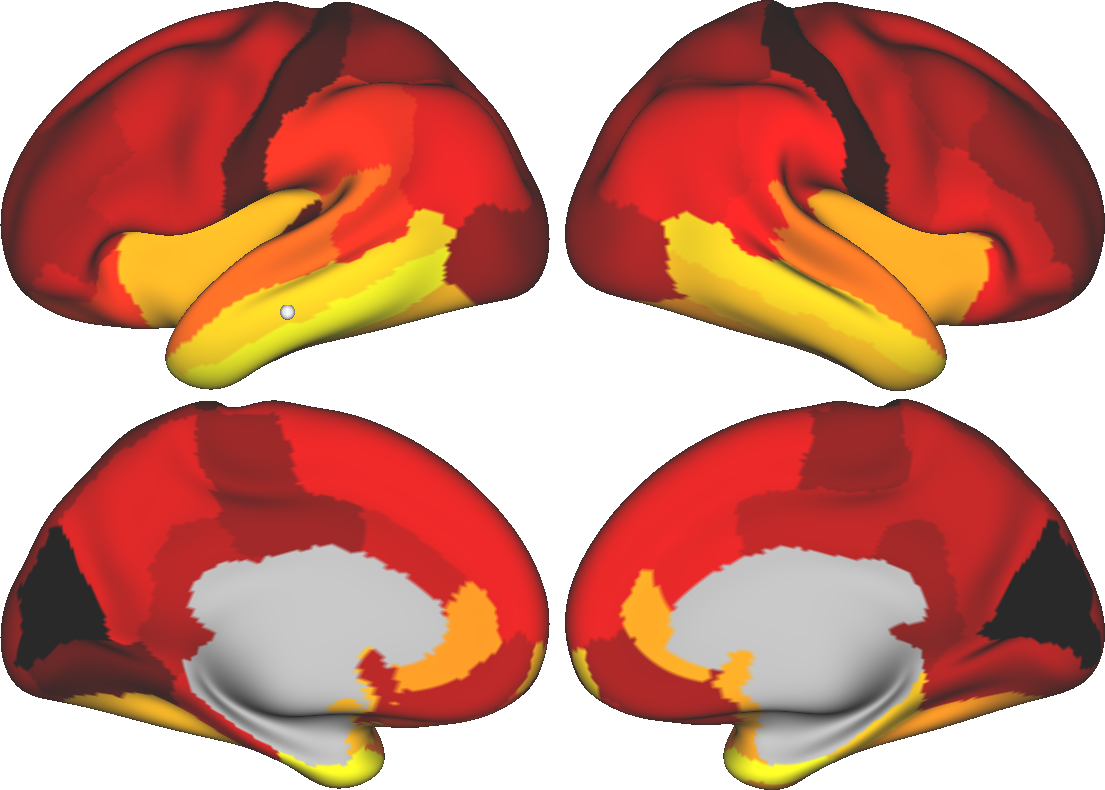} \\
            
            \hline
            \\
            & \multicolumn{2}{c}{\includegraphics[width=0.2\textwidth,valign=c]{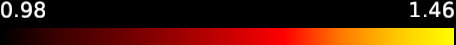}} 
            & \multicolumn{2}{c}{\includegraphics[width=0.2\textwidth,valign=c]{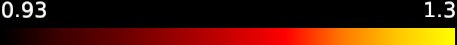}} 
            & \multicolumn{2}{c}{\includegraphics[width=0.2\textwidth,valign=c]{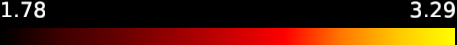}} \\
        \end{tabular}
        \caption*{\textbf{Panel C:} SUVR maps of biomarker comparisons between LENO-simulated and true data for $A$, $\tau$, and $N$ during the \textbf{training} period.}
    \end{subfigure}
    \hfill

    \begin{subfigure}[!htbp]{1\textwidth}
        \centering
        \begin{tabular}{ccc||cc||cc}
            \hline
            Biomarker
            &\multicolumn{2}{|c||}{\textbf{$A$}} &\multicolumn{2}{c||}{\textbf{$\tau$}}
            &\multicolumn{2}{c}{\textbf{$N$}}\\ 
            \hline
            \multicolumn{1}{c|}{Age} & 74 & 77 & 74 & 80 & 75 & 81\\
            \hline
            \multicolumn{1}{c|}{LENO}
            &\includegraphics[width=0.12\textwidth,valign=c]{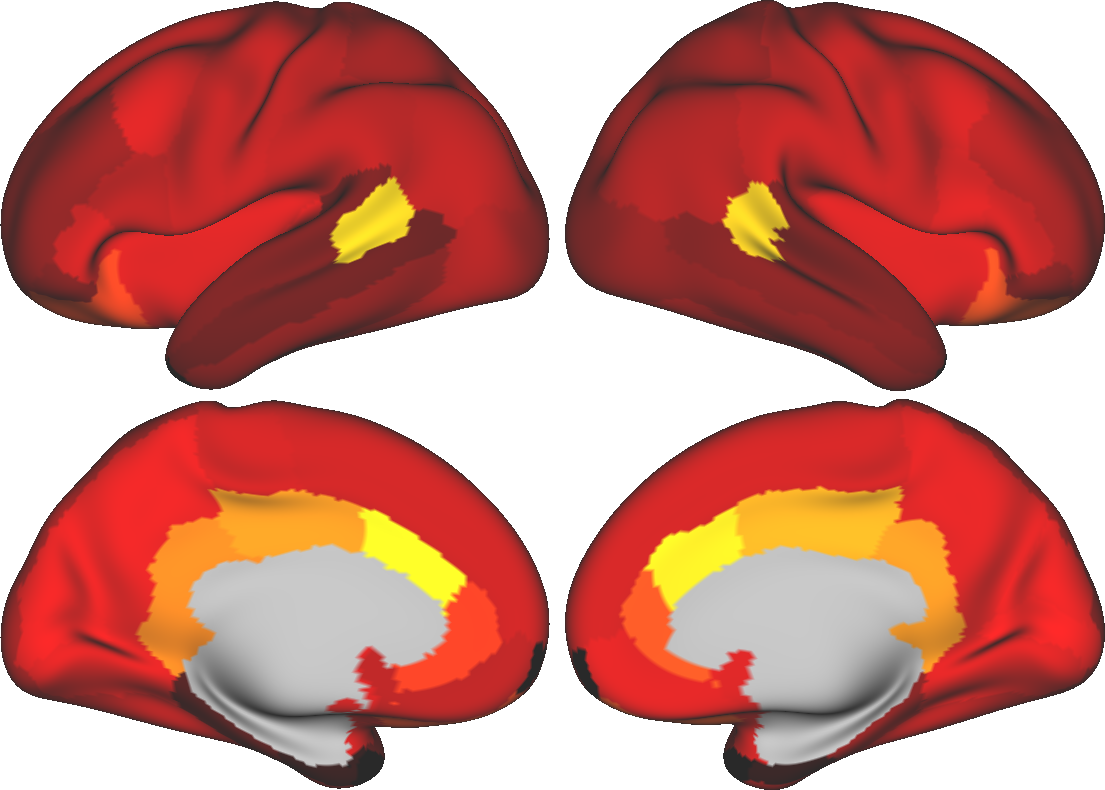} &
            \includegraphics[width=0.12\textwidth,valign=c]{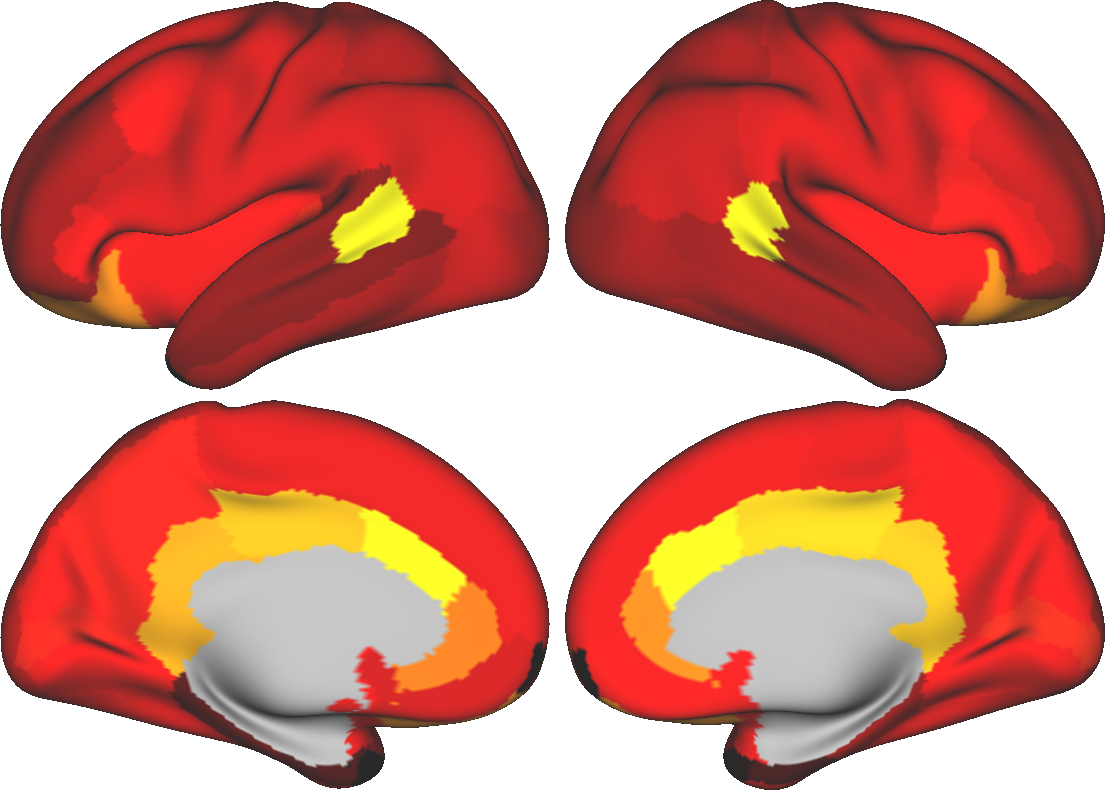} &
            \includegraphics[width=0.12\textwidth,valign=c]{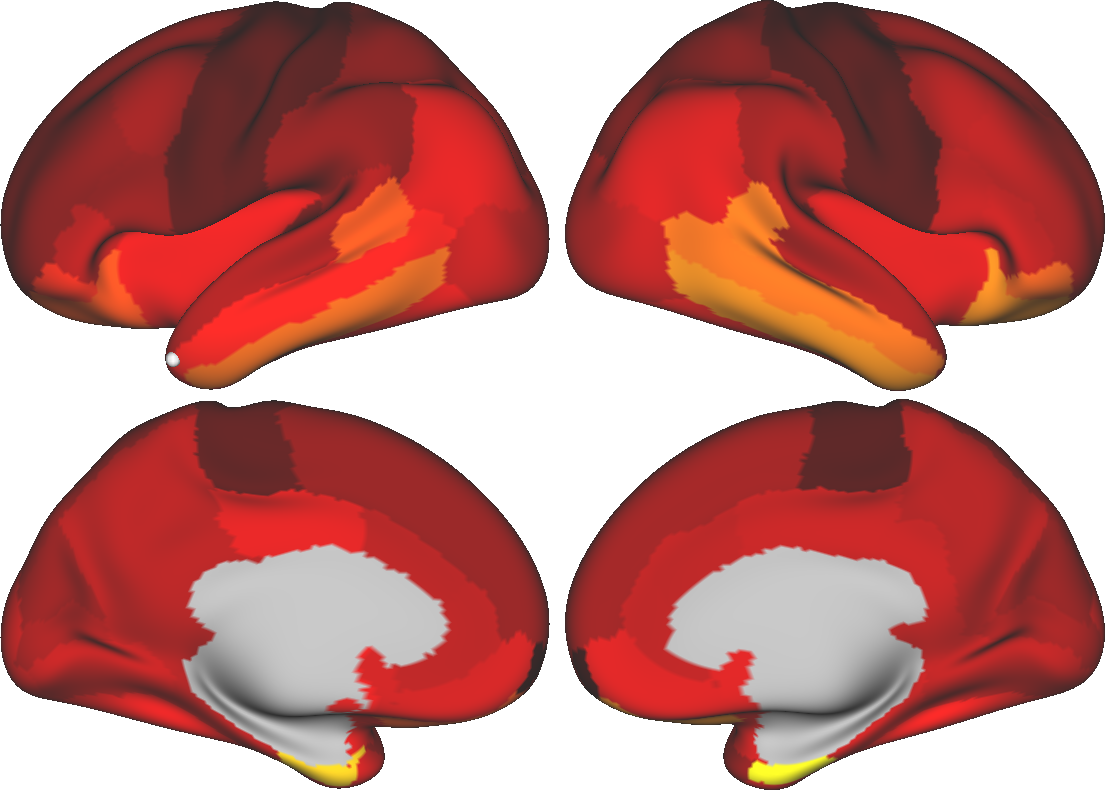} &
            \includegraphics[width=0.12\textwidth,valign=c]{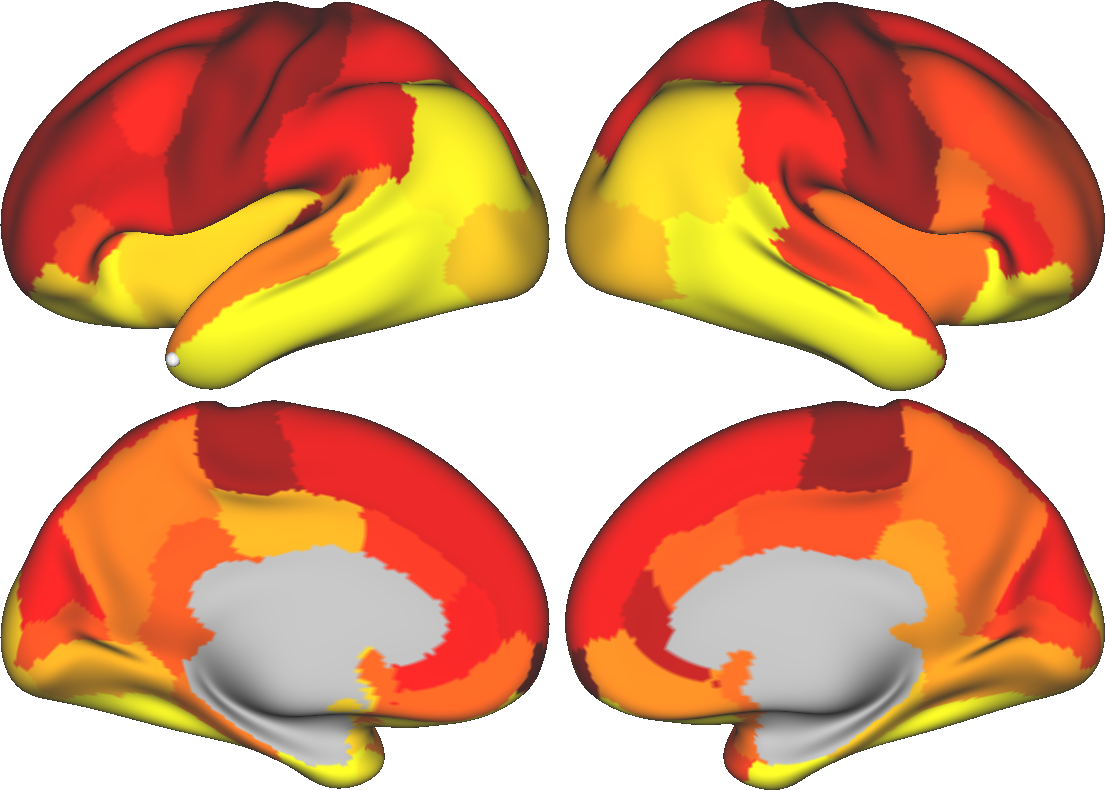} &
            \includegraphics[width=0.12\textwidth,valign=c]{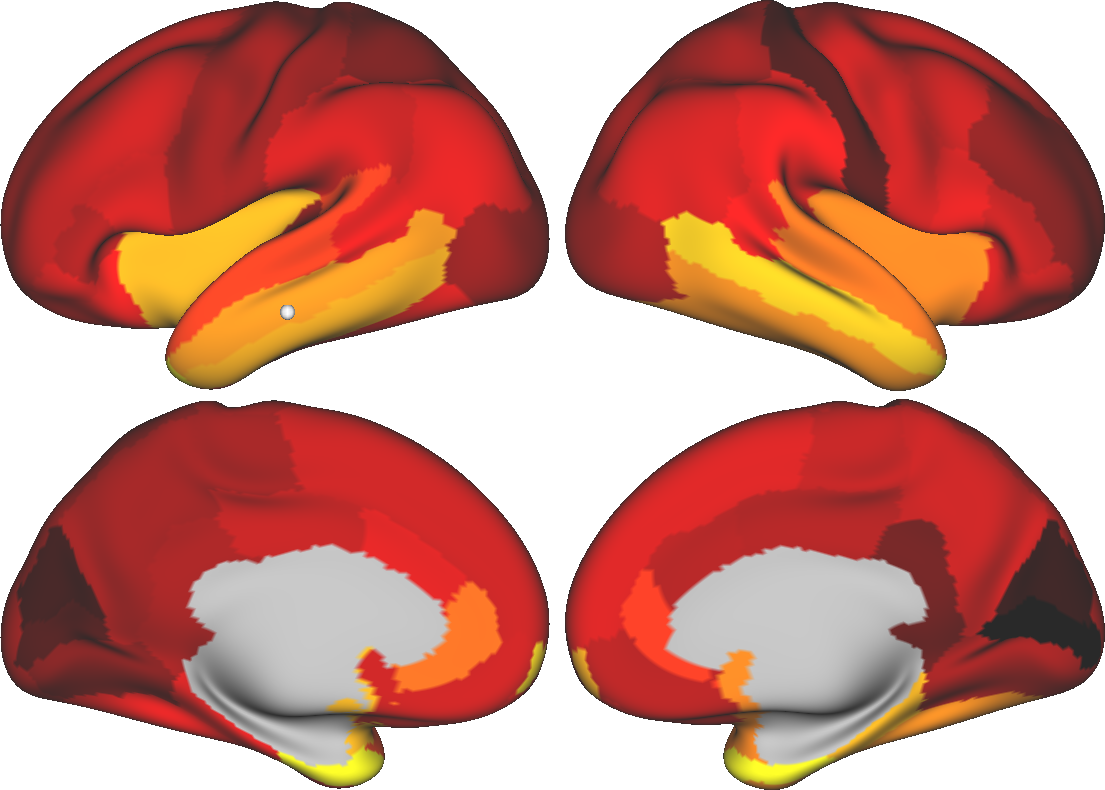} &
            \includegraphics[width=0.12\textwidth,valign=c]{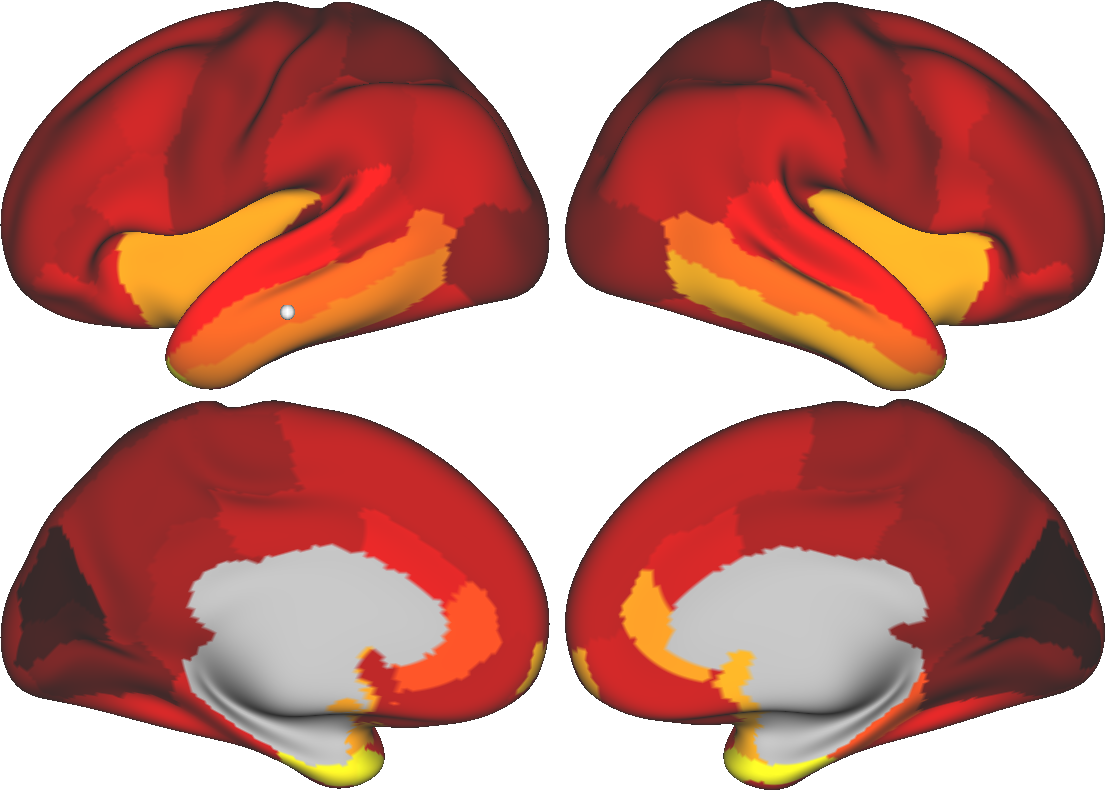} \\
            \hline
            \multicolumn{1}{c|}{Truth} &\includegraphics[width=0.12\textwidth,valign=c]{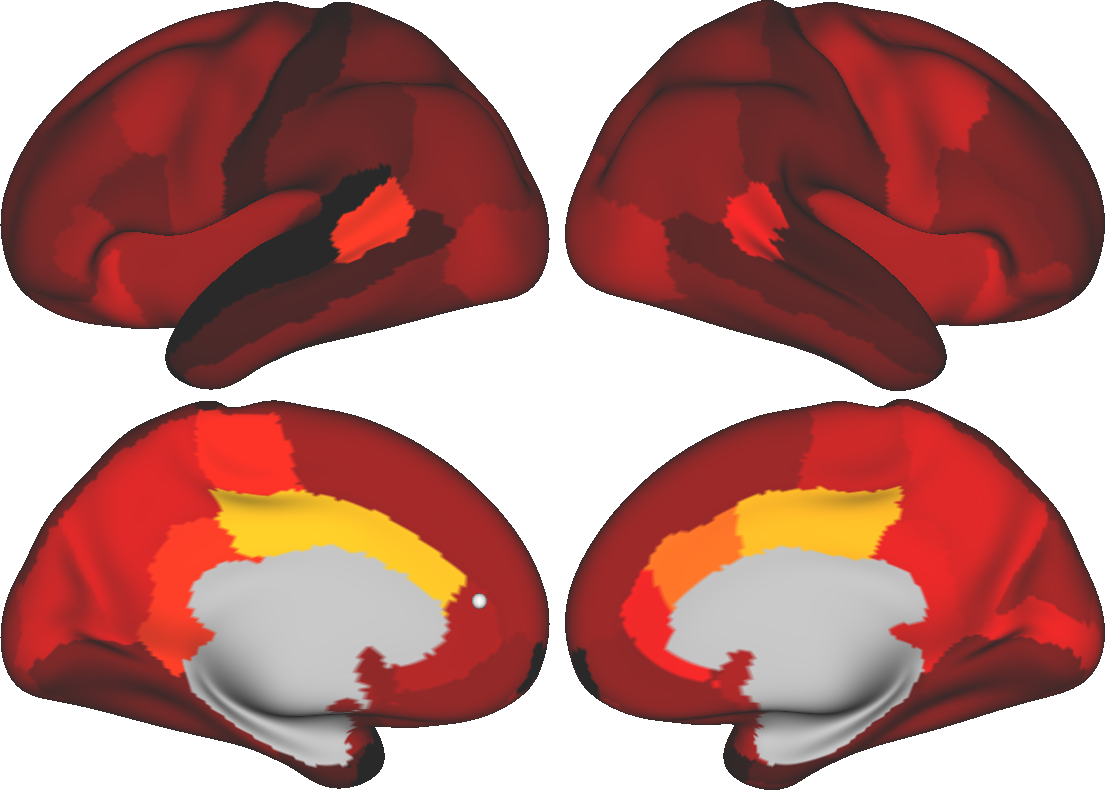} &
            - &
            \includegraphics[width=0.12\textwidth,valign=c]{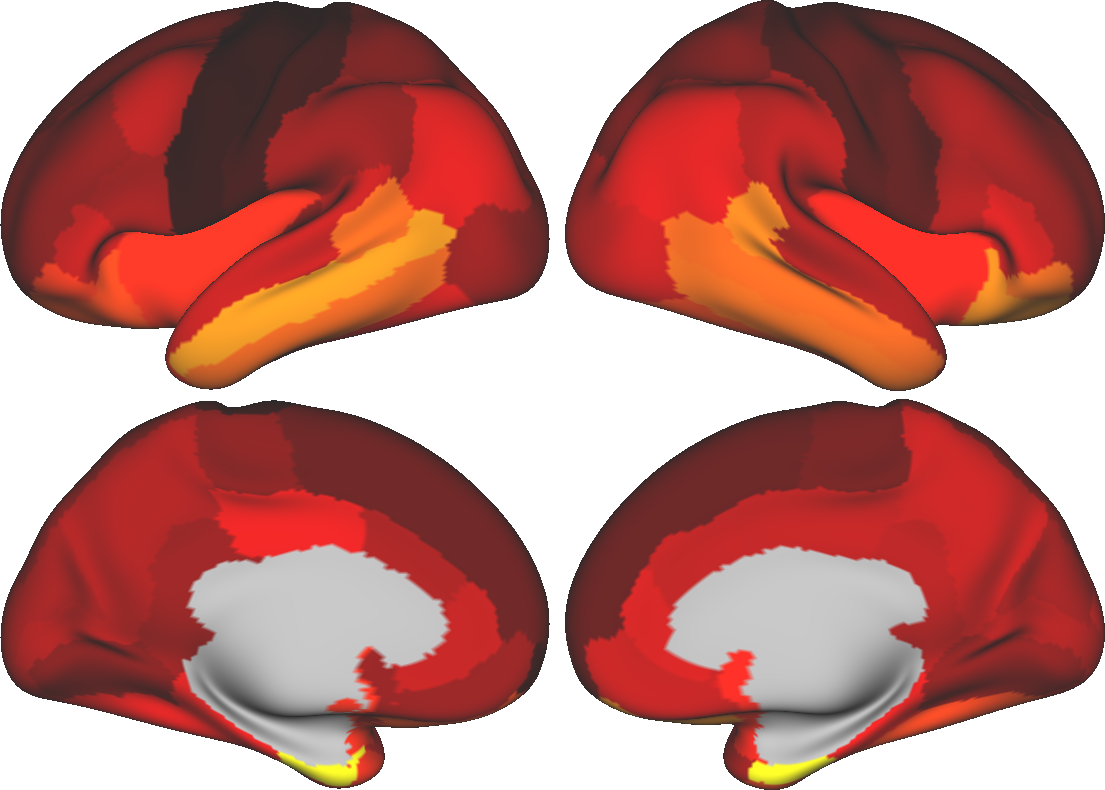} &
            - &
            \includegraphics[width=0.12\textwidth,valign=c]{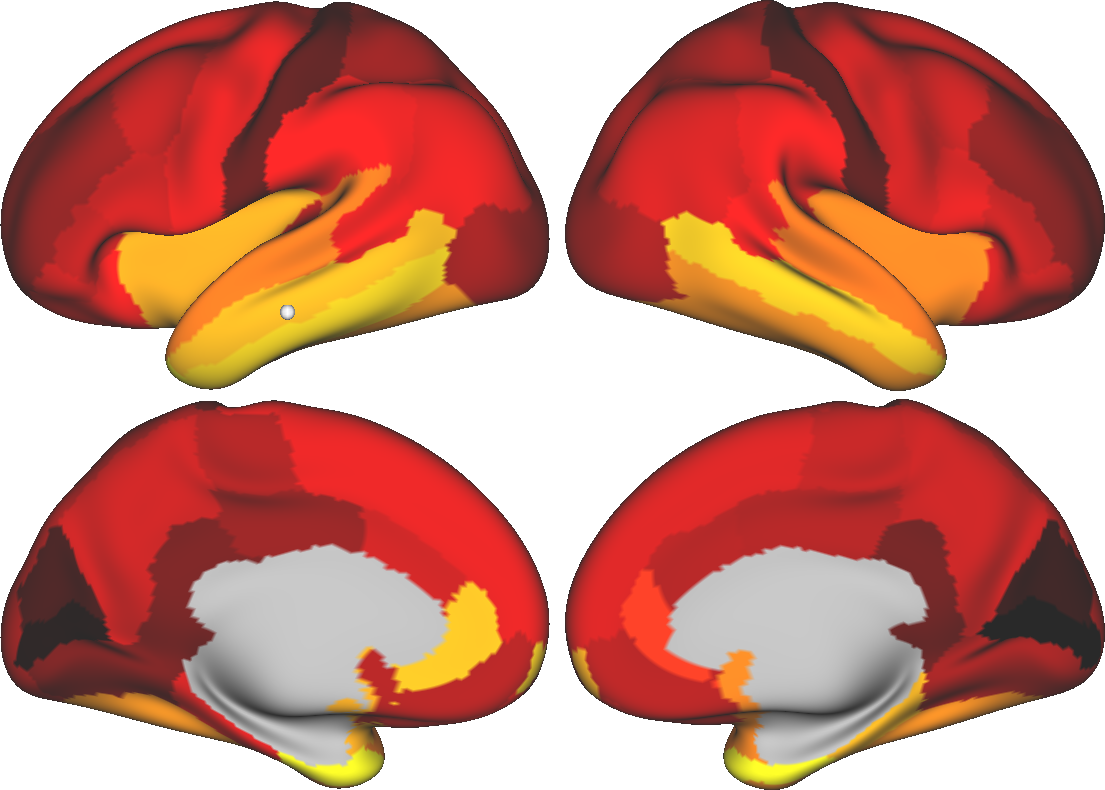} &
            - \\
            
            \hline
            \\
            & \multicolumn{2}{c}{\includegraphics[width=0.2\textwidth,valign=c]{figures/abeta/abeta-bar.png}} 
            & \multicolumn{2}{c}{\includegraphics[width=0.2\textwidth,valign=c]{figures/tau/tau-bar.png}} 
            & \multicolumn{2}{c}{\includegraphics[width=0.2\textwidth,valign=c]{figures/n/n-bar.png}} \\
        \end{tabular}
        \caption*{\textbf{Panel D:} SUVR maps of biomarker comparisons between LENO-simulated and true data for $A$, $\tau$, and $N$ during the \textbf{prediction} period.}
    \end{subfigure}

\caption{\textbf{Panel A:} Image preprocessing pipeline for constructing the graph Laplacian matrix $L$. \textbf{Panel B:} First four eigenfunctions of the graph Laplacian in the 3D brain domain. \textbf{Panel C:} SUVR maps of biomarker comparisons for $A$, $\tau$, and $N$ during the training period.
\textbf{Panel D:} SUVR maps of biomarker comparisons $A$, $\tau$, and $N$ during the prediction period. In Panels C and D, the top row displays SUVR distributions generated by the model at selected ages, while the bottom row shows the corresponding ground truth measurements. Color bars indicate SUVR intensity across brain regions.}
    \label{fig:atn}
\end{figure}

{\bf Visualization of Temporal Predictions:}  
Figure~\ref{fig:atn} visualizes the predicted and ground truth SUVR maps for $A$, $\tau$, and $N$ over time. In each panel, the top row displays the model-predicted SUVR distributions at selected ages during both the training and prediction intervals, while the bottom row presents the corresponding measured SUVR distributions. Color bars indicate SUVR intensity across brain regions.

This figure illustrates the temporal progression of the three key AD imaging biomarkers in a representative patient, showing accurate reproduction of both spatial patterns and temporal growth trends within and beyond the training window. It highlights the learned and predicted dynamics of tau accumulation and neurodegeneration, demonstrating the model’s ability to generalize complex spatiotemporal biomarker dynamics from limited longitudinal data within anatomically constrained brain networks.

\begin{figure}[!htbp]
	\centering
	\begin{tabular}{ccc}
		 \textbf{CN} & \textbf{MCI} & \textbf{AD} \\
		 \includegraphics[width=0.3\textwidth]{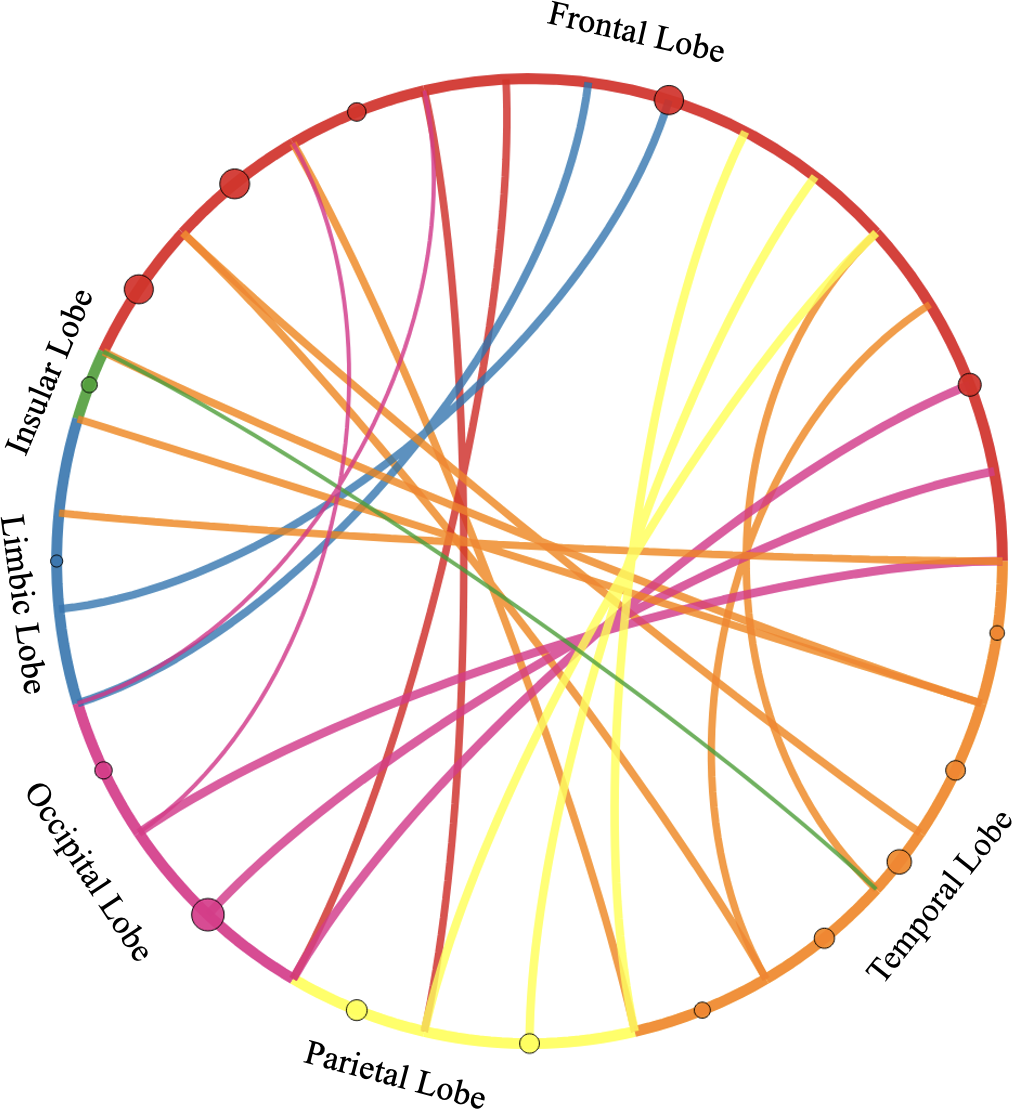} & \includegraphics[width=0.3\textwidth]{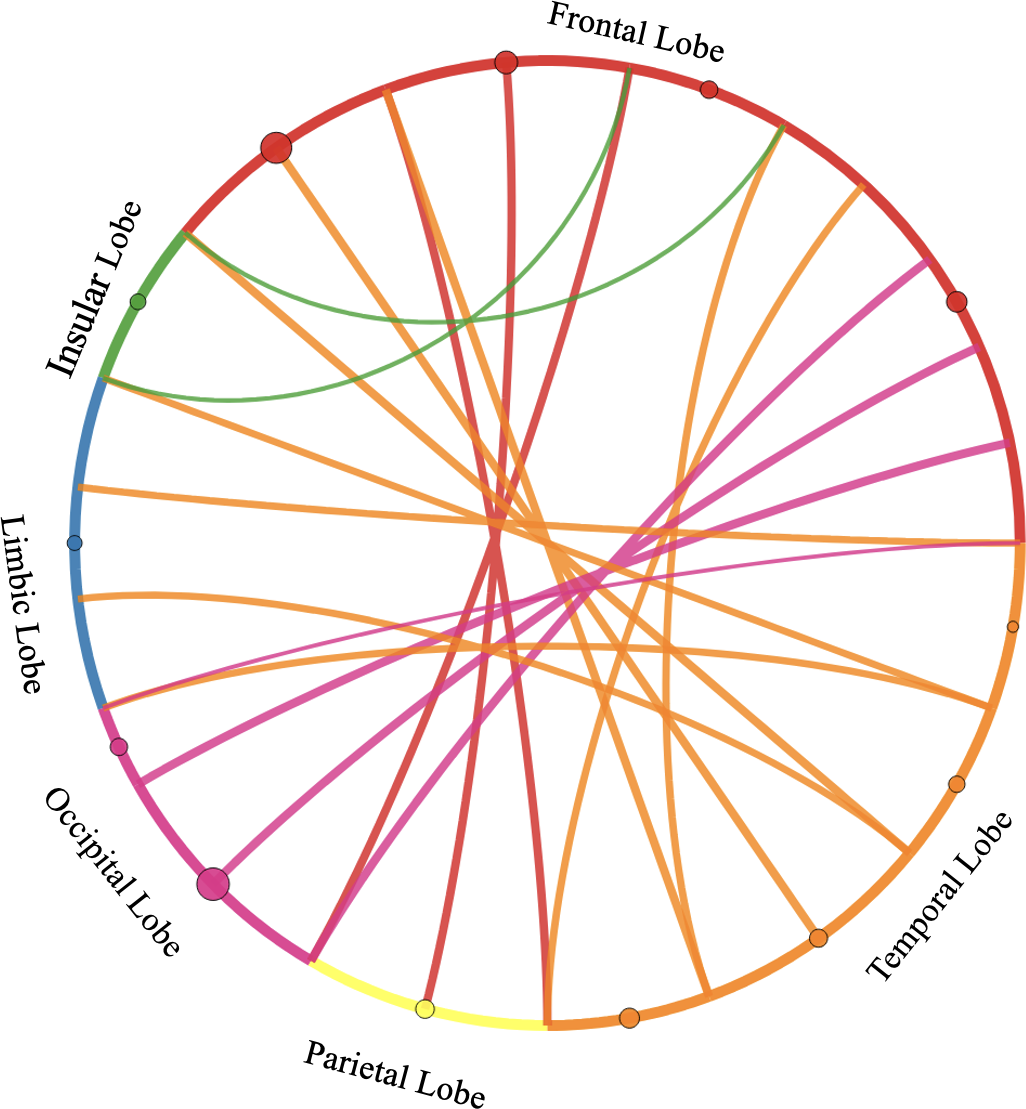} & \includegraphics[width=0.3\textwidth]{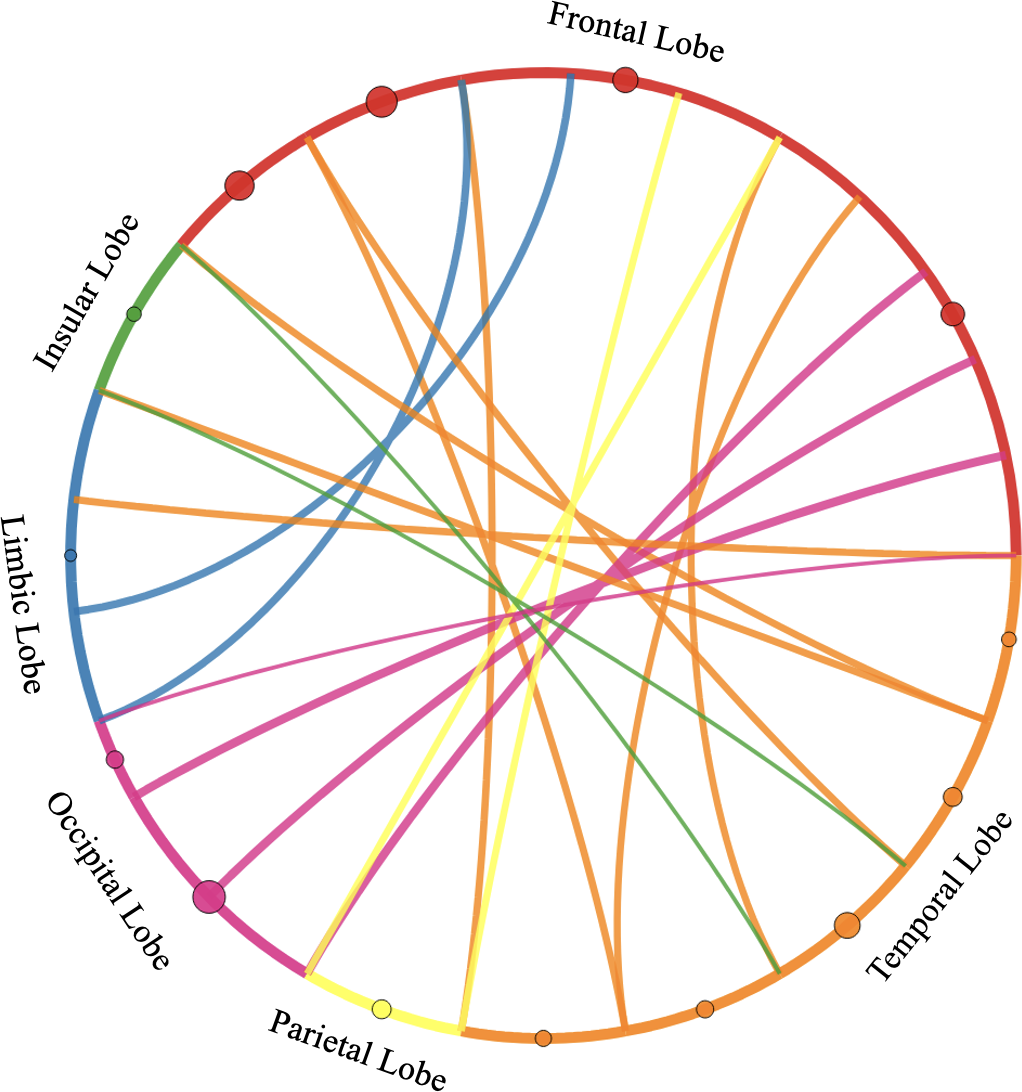} \\
        \multicolumn{3}{c}{Chord diagrams of $A$}\\
		 \includegraphics[width=0.3\textwidth]{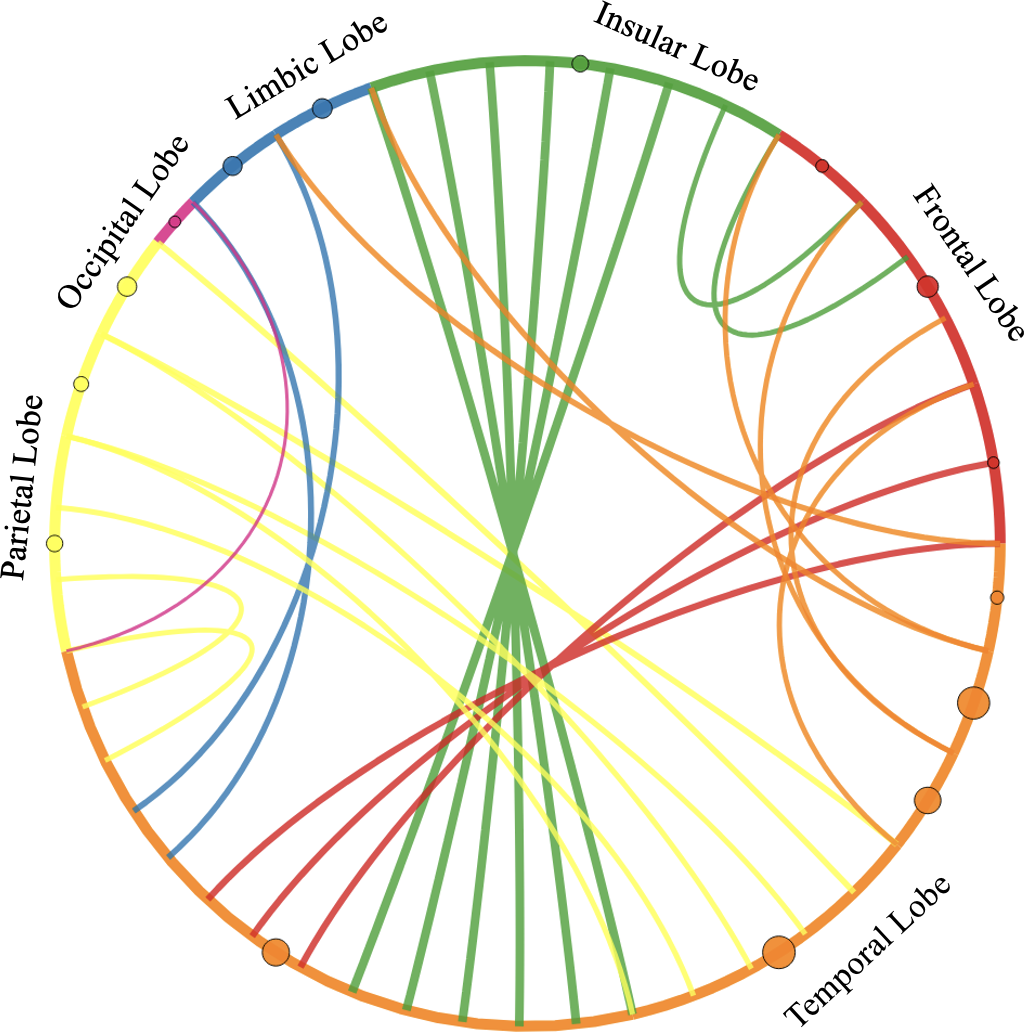}   & \includegraphics[width=0.3\textwidth]{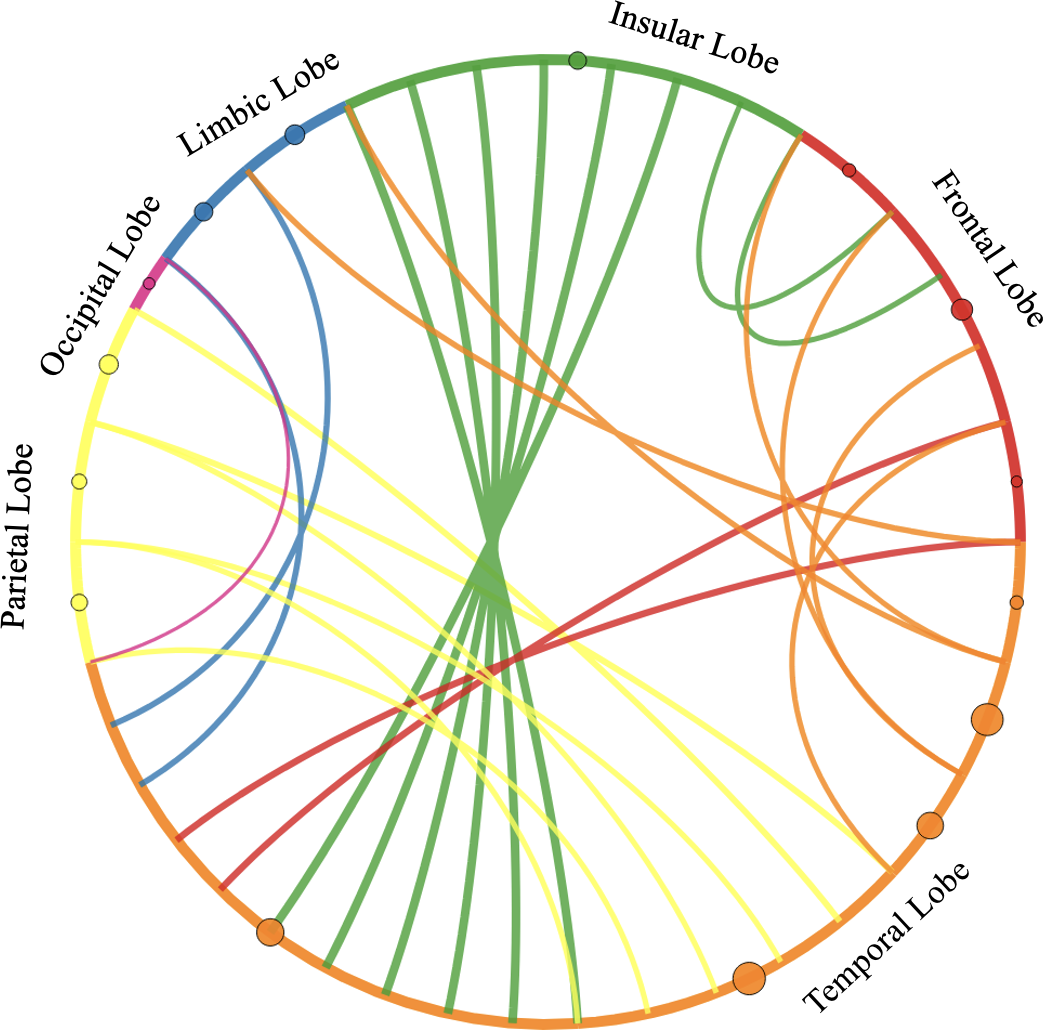}   & \includegraphics[width=0.3\textwidth]{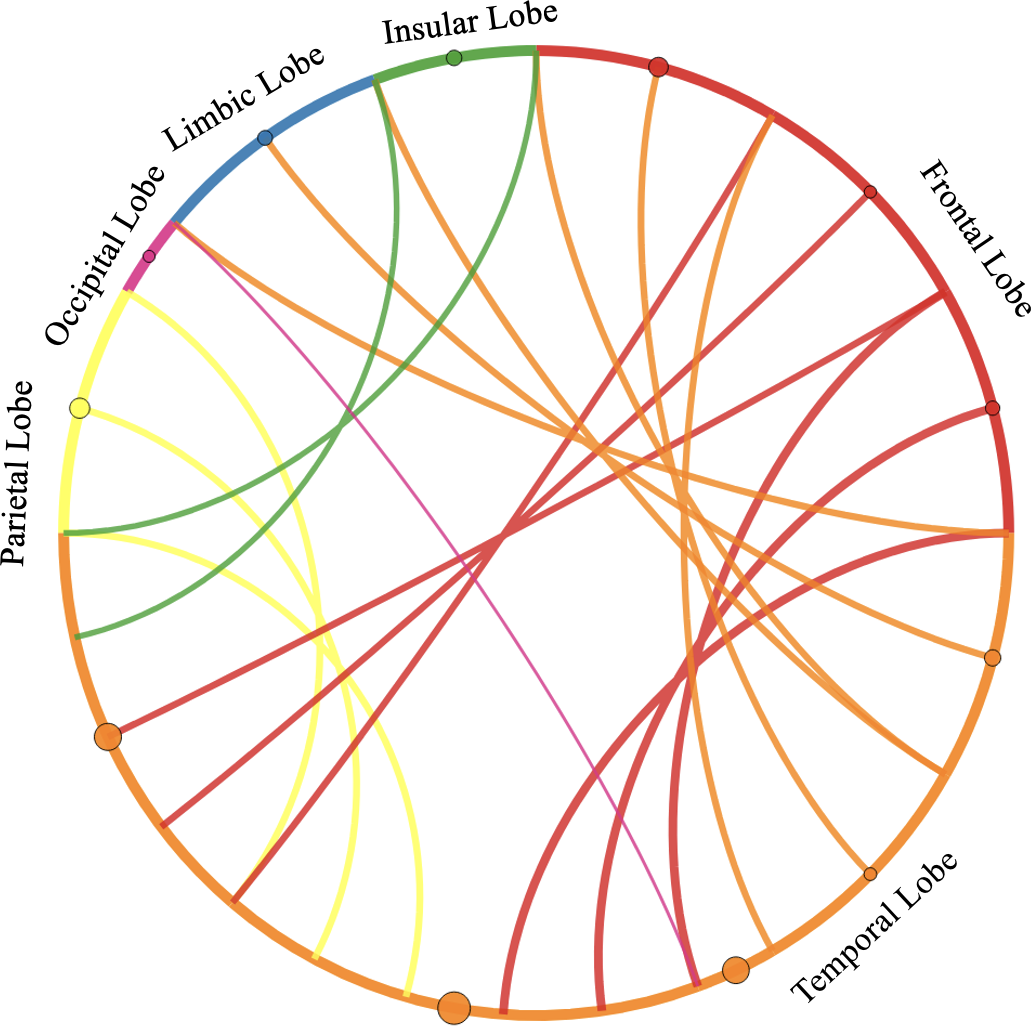} \\
         \multicolumn{3}{c}{Chord diagrams of $\tau$}\\
		\includegraphics[width=0.3\textwidth]{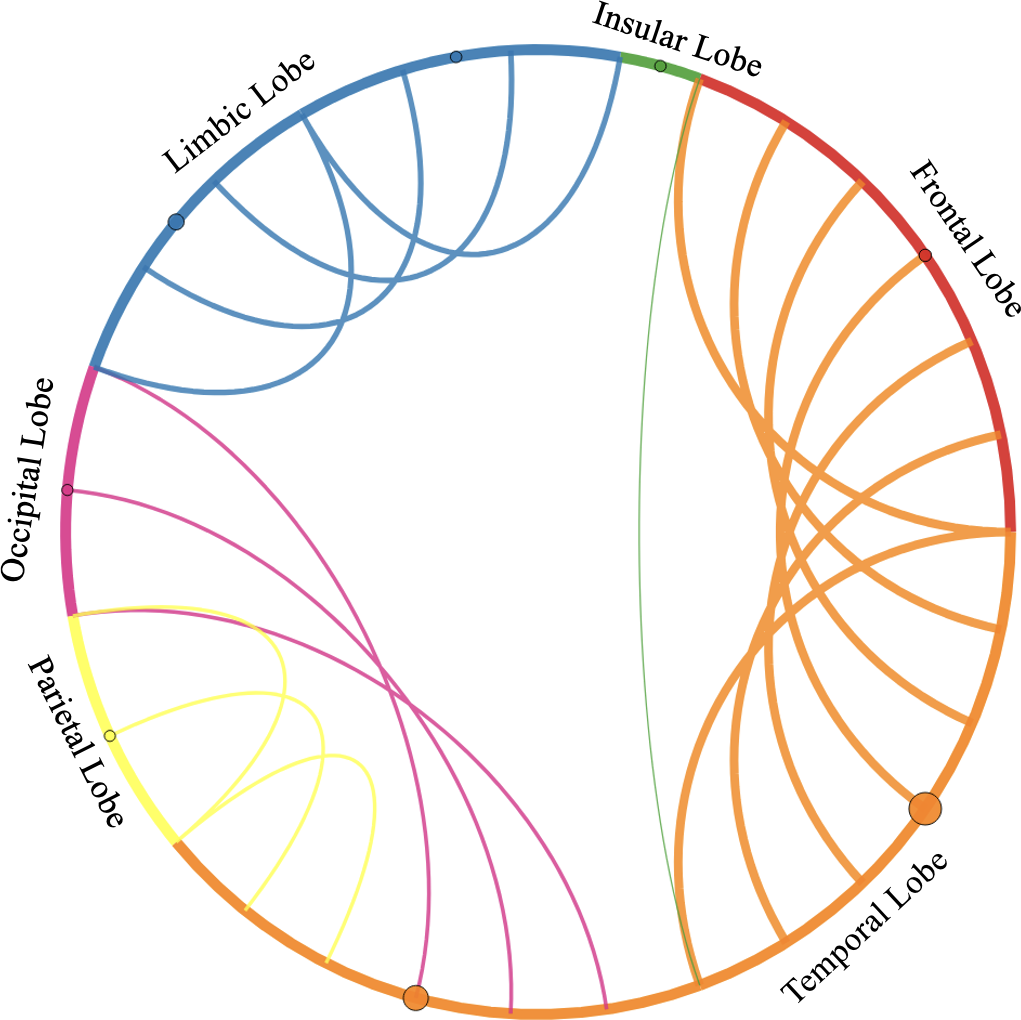}     & \includegraphics[width=0.3\textwidth]{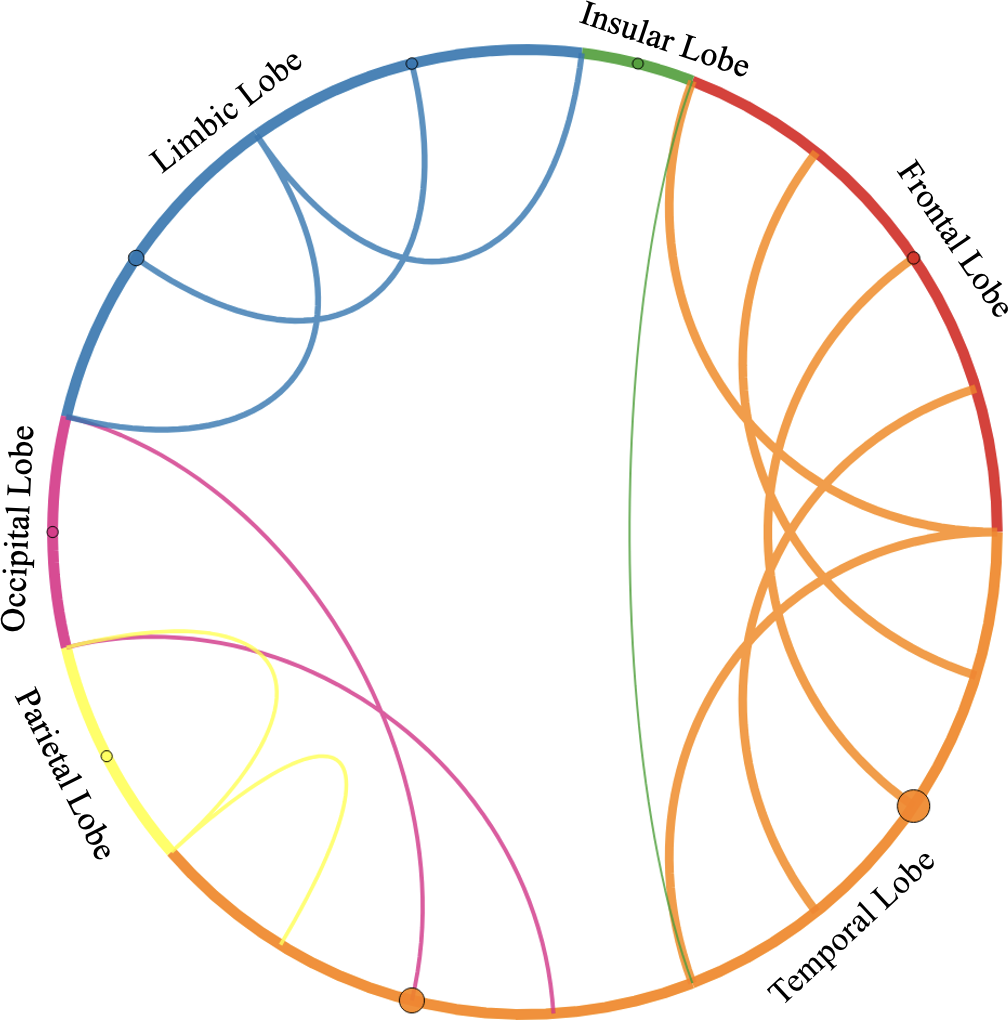}     & \includegraphics[width=0.3\textwidth]{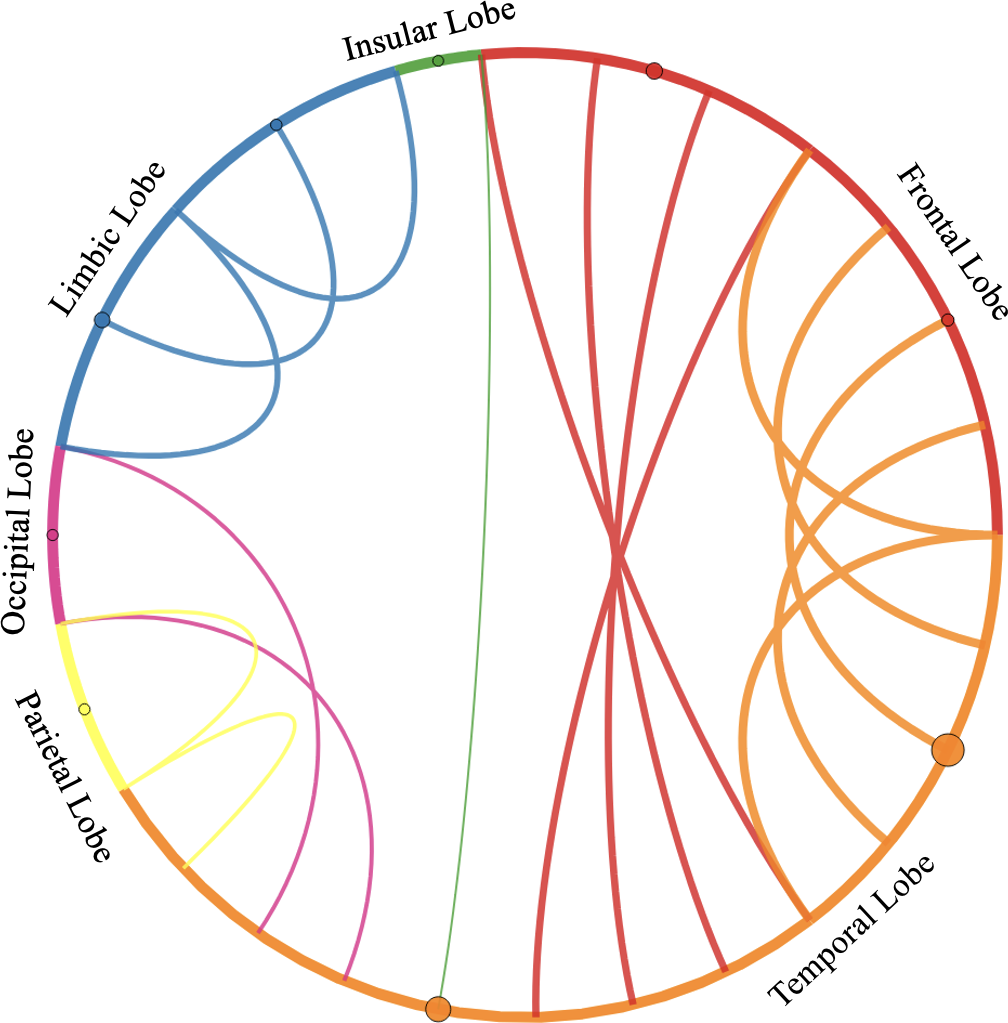} \\
        \multicolumn{3}{c}{Chord diagrams of $N$}\\
	\end{tabular}
	\caption{Chord diagrams illustrating inter-regional SUVR connectivity patterns for A$\beta$, $\tau$, and neurodegeneration biomarkers across cognitive stages. Each row represents one biomarker and each column corresponds to a diagnostic group: CN, MCI, and AD. The thickness and color of the chords indicate the strength and nature of model-inferred connections between brain regions based on SUVR distributions. As disease progresses from CN to AD, inter-regional connectivity becomes increasingly localized and asymmetric, particularly for tau and neurodegeneration, reflecting stage-dependent disruptions in network architecture associated with Alzheimer’s progression.}
	\label{fig:chrod}
\end{figure}

{\bf Connectivity Patterns Across Disease Stages:} To uncover structural differences in biomarker propagation across cognitive stages (CN, MCI, AD), we visualize chord diagrams for $A$, $\tau$, and $N$ in Figure~\ref{fig:chrod}. These diagrams illustrate the inter-regional connectivity strengths, derived from the model-predicted SUVR dynamics, capturing how the spatial distribution and coupling of pathology evolve across disease progression.
In CN subjects, connectivity was sparse and largely localized to adjacent regions. In the MCI stage, both amyloid and tau exhibited increased coupling, particularly involving temporal, insular, and parietal lobes, with tau showing notably stronger and broader propagation patterns. By the AD stage, amyloid connectivity became widespread across frontal, temporal, and parietal regions, while tau displayed dense and extensive inter-regional connections, consistent with advanced Braak staging. Neurodegeneration patterns closely followed tau propagation, showing progressive strengthening and expansion of connections into temporal, insular, and parietal regions. These results demonstrate the model’s ability to capture distinct, biomarker-specific spatiotemporal propagation signatures across disease stages, faithfully reflecting known anatomical trajectories of Alzheimer’s pathology.

\textbf{Modeling Cognitive Score Dynamics:} Beyond molecular pathology, we further investigate the dynamics of cognitive decline through a scalar variable \( C \), representing cognitive scores (e.g., MMSE), governed by
\begin{equation}
	\begin{cases}
		C_t = \mathcal{F}_4(N), & t \in (0,T), \\
		C(0) = c_0,
	\end{cases}
	\label{eq:c}
\end{equation}
where the rate of cognitive decline depends on the neurodegeneration profile \( N \). Here, the nonlinear functional \(\mathcal{F}_4: \mathbb{R}^n \to \mathbb{R}\) maps the vector-valued \( N \) (derived from SUVR data) to the scalar cognitive score derivative. To approximate \(\mathcal{F}_4\), we employ a neural network \(\mathcal{N}\) that takes the full neurodegeneration profile \( N \) as input.
Our model achieves strong accuracy in both training and prediction of cognitive trajectories. Table~\ref{tab:combined_accuracy} summarizes the \( L^2 \) accuracy metrics.


\subsection{Effect of Personalized Treatment Strategies}

\begin{figure}[!htbp]
	\centering
	\includegraphics[width=0.95\textwidth]{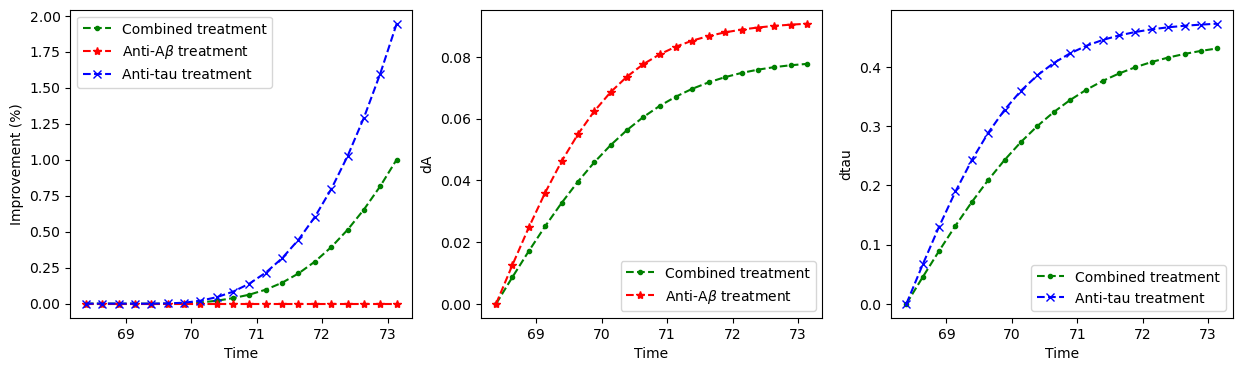}
\caption{Left: Simulated trajectories of cognitive scores under three treatment strategies—combined (Anti-A$\beta$ + Anti-tau), Anti-A$\beta$ alone, and Anti-tau alone. Middle: dose regimen of Anti-A$\beta$ (both single and combined). Right: dose regimen of Anti-tau (both single and combined).}
	\label{fig:treat_effect}
\end{figure}

To simulate the impact of hypothetical interventions, we introduce treatment terms that attenuate the accumulation of \( A \) and \(\tau\):
\begin{equation}
	\begin{cases}
		A_t - \alpha_A L A = \mathcal{F}_1(A) - d_A A, \\
		\tau_t - \alpha_\tau L \tau = \mathcal{F}_2(A, \tau) - d_\tau \tau, \\
	\end{cases}
	\label{eq:abetanbrain}
\end{equation}
with initial conditions as previously specified. The terms \( d_A A \) and \( d_\tau \tau \) represent the effects of anti-amyloid and anti-tau treatments, respectively. We parameterize the treatment intensities \( d_A \) and \( d_\tau \) using sigmoid neural networks to enforce biologically plausible upper-bound constraints that vary with time \( t \).
To explore optimal treatment strategies, these neural networks are trained to minimize a cognitive decline loss function subject to treatment constraints, as described in the \textbf{Methods} section.
Figure~\ref{fig:treat_effect} presents the predicted cognitive score trajectories under various treatment regimens. The results suggest that anti-\(\tau\) treatment produces more consistent cognitive improvement compared to anti-\(A\) treatment.

\section{Discussion}
In this work, we developed an innovative PDE learning framework that directly infers the spatiotemporal dynamics governing AD biomarkers from longitudinal neuroimaging data. Leveraging the LENO, our approach integrates mechanistic insights from brain network topology with data-driven nonlinear operator learning to model the propagation of amyloid-$\beta$ ($A$), tau ($\tau$), and neurodegeneration ($N$) across personalized anatomical connectivity graphs. This fusion of interpretable mathematical modeling and advanced machine learning represents a powerful advancement for characterizing the multiscale, heterogeneous nature of AD progression.

A key strength of our framework is its foundation in interpretable mathematical modeling through the use of Laplacian eigenfunctions as spatial basis functions. The Laplacian eigenfunctions, intrinsic to the underlying brain geometry or network topology, provide a natural and physically meaningful decomposition of spatial patterns. Each eigenfunction corresponds to a spatial mode with an associated frequency, enabling a hierarchical representation of biomarker distributions that aligns with known neuroanatomical and functional gradients. This spectral decomposition not only facilitates efficient numerical approximation on irregular brain geometries but also lends interpretability to the learned operators by revealing how distinct spatial modes contribute to disease propagation. Such an approach contrasts with black-box machine learning models, offering transparency and mechanistic insight that are essential for clinical trust and hypothesis generation.

The clinical and scientific significance of this framework lies in its capacity to learn interpretable, biologically grounded PDE operators that describe disease dynamics in irregular brain geometries, enabling robust forward prediction and patient-specific adaptation through transfer learning. Traditional models of AD progression have relied on either simplified compartmental approaches or purely statistical predictions, often lacking spatial resolution or mechanistic interpretability. Our model overcomes these limitations by directly incorporating anatomical connectivity, which shapes biomarker diffusion and accumulation, thus reflecting the true biological substrate of neurodegenerative spread.

The stage-dependent connectivity patterns revealed by our model align closely with established theories of AD propagation, notably the network diffusion hypothesis proposed by Raj et al.~\cite{raj2012network}. Specifically, our inferred propagation pathways capture the well-characterized transition of tau pathology from localized regions in CN individuals to extensive limbic and neocortical involvement in AD, mirroring neuropathological Braak staging. This spatial progression is accompanied by increasingly complex coupling between amyloid, tau, and neurodegeneration, particularly within temporoparietal and insular cortices, consistent with findings from longitudinal multimodal imaging studies~\cite{iturria2016early, jagust2018imaging}. Our ability to reconstruct these detailed biomarker interdependencies and their evolving network topology from limited patient-specific SUVR data underscores the potential of data-driven PDE learning to capture multiscale neurodegenerative processes in a personalized manner.

Moreover, the integration of cognitive decline dynamics as a downstream outcome of neurodegeneration bridges molecular pathology with clinical presentation. The model’s high accuracy in predicting cognitive trajectories directly from imaging biomarkers demonstrates its translational utility, offering a quantitative framework for early diagnosis, prognosis, and patient stratification. By linking network-level biomarker evolution to cognitive scores, our framework enables a more holistic understanding of disease progression that transcends static biomarkers alone.

Critically, the introduction of treatment effect terms within the PDE framework allows simulation of personalized therapeutic interventions, offering an \textit{in silico} platform for exploring the differential impacts of anti-amyloid and anti-tau therapies on cognitive outcomes. Our findings that anti-tau interventions yield more consistent cognitive improvements corroborate emerging clinical evidence and highlight the utility of mechanistic models for guiding therapy optimization. This capability could accelerate the development and refinement of targeted treatments by predicting patient-specific responses based on their unique biomarker and network profiles.

From a methodological perspective, the LENO framework’s reliance on Laplacian eigenfunctions as spatial basis functions facilitates efficient learning on irregular brain geometries and diverse network topologies. The demonstrated robustness to geometric variation and heterogeneity across patients, enabled by transfer learning through personalized temporal scaling, provides a scalable approach to modeling AD across populations. This adaptability is crucial for capturing the complex inter-individual variability that challenges conventional modeling techniques.

In conclusion, our work establishes a versatile, interpretable, and personalized PDE-based modeling paradigm that unites clinical imaging data, network neuroscience, and nonlinear dynamics. This framework not only advances fundamental understanding of AD pathology and its network-driven spread but also lays the groundwork for digital twin technologies in neurodegeneration—integrated computational models capable of predicting individual disease trajectories and optimizing therapeutic strategies. As such, it holds promise for transforming both research and clinical management of AD.

\section{Methods}
This section presents an overview of the proposed Laplacian Eigenfunction Neural Operator (LENO) framework, illustrated in Figure~\ref{fig:nn} and detailed in~\cite{hao2025laplacian}.

We consider a general nonlinear reaction-diffusion system of the form
\begin{equation}\label{eq:baseeq}
	\left\{
	\begin{aligned}
		u_t-\alpha\Delta u&= \mathcal{F}(u), & \text{ in }\Omega\times [0,T]\\
		u(x,0) &= u_0(x), &\text{ in }\Omega,
	\end{aligned}\right.
\end{equation}
where \( u \) represents a biomarker vector (e.g., A$\beta$, tau, or neurodegeneration) evolving within the brain geometry \( \Omega \subset \mathbb{R}^d \), \( \alpha \) is a diffusion coefficient, and \( \mathcal{F} \) is an unknown nonlinear reaction term to be inferred from empirical data. Non-flux boundary conditions are imposed. The weak form reads: Find $u\in L^2(0,T;H^1(\Omega))$ with $u_t\in L^2(0,T;H^{-1}(\Omega))$ such that
\begin{equation}\label{eq:originvp}
	(u_t,v)_{L^2}+(\alpha\nabla u,\nabla v)_{L^2} = (\mathcal{F}(u),v)_{L^2} 
\end{equation}
for $ v \in H^1(\Omega)$ with $u(x,0) = u_0(x)$.
The nonlinear term  $\mathcal{F}$  is treated as a nonlinear operator:
\begin{equation}
	\mathcal{F}:H^1(\Omega) \rightarrow H^{-1}(\Omega). 
\end{equation} 

To enable data-driven discovery of $\mathcal{F}$, we approximate it using a neural operator $\mathcal{N}(u;\theta)$, parameterized by weights $\theta$. Following~\cite{hao2025laplacian}, the operator is constructed using Laplacian eigenfunctions as basis functions, providing an efficient and geometry-aware representation ideal for structured and unstructured domains.
This approach not only delivers highly accurate predictions of system dynamics but also offers mechanistic insight into the causal structure underlying the learned dynamics. Moreover, when applied to human brain domains, the model seamlessly incorporates finite element methods to compute Laplacian eigenfunctions~\cite{susanne1994mathematical,boffi2010finite}, ensuring both numerical stability and geometric flexibility.

{\bf Eigenfunction-Based Neural Operator:}
For a general positive definite operator \( L \) (e.g., \( -\Delta \)), its eigenfunctions provide a basis for representing the solution. Given Hilbert spaces \( \mathcal{X}(\Omega) \) and \( \mathcal{Y}(\Omega) \), we define:
\begin{equation}
    \mathcal{N}(u) = \sum_{i=1}^P A_i\, \sigma(W_i u + B_i)\, \phi_i(x),
\end{equation}
where \( A_i \), \( W_i \), and \( B_i \) are trainable operators, and \( \sigma \) is a nonlinear activation function. Unlike conventional neural networks, Laplacian eigenfunctions \( \{\phi_i\} \) naturally capture spatial structure.

{\bf Learning Problem:}
For the learning problem, we will project both the data and the equation onto the space $	V_P = {\rm span}\{\phi_i\}_{i=1}^P.$

{\bf Data $L^2$ projection.}  Given the observed data consisting of data samples $\{u^n(x), t^n\}_{n=0}^N$ at time $t_n$, the projection is defined as: $\beta_i^n=(u^n,\phi_i)_{L^2}$, $i=1,\cdots,P$.

{\bf Equation projection.} By restricting the variational problem \eqref{eq:originvp} to the subspace $V_P$, we approximate the numerical solution as 
$\displaystyle \widetilde{u}_P^n(x)=\sum_{i=1}^{P}\widetilde{\beta}_i^n\phi_i(x).$

Using the semi-implicit Euler method for time discretization and replacing  $\mathcal{F}$  with  $\mathcal{N}$, we derive the following discrete formulation
\begin{equation}\label{eq:discrete}
	\frac{\widetilde{\bm{\beta}}^n-\widetilde{\bm{\beta}}^{n-1}}{ t_n-t_{n-1}}+D\Lambda_P\widetilde{\bm{\beta}}^n = \mathcal{G}(\widetilde{\bm{\beta}}^{n-1};\theta), \quad  n=1,\cdots,N,
\end{equation}
where $\bm{\widetilde{\beta}}^n=(\widetilde{\beta}_1^{n},\cdots,\widetilde{\beta}_P^{n})^T$ and $\Lambda_P = {\rm diag}(\lambda_1,\cdots,\lambda_P)$ (due to the orthogonality of $\phi_i$). The nonlinear component of the neural operator, originally expressed as ${A}_i\sigma(\mathcal{W}_i{u}+\mathcal{B}_i)$ is now reduced to a continuous function $\mathcal{G}_i(\tilde{\beta};\theta)$ given by neural network, which depends on the numerical approximation $\tilde{\bm \beta}$ of solution coefficient on the eigenfunction space.  We denote the collection as $\mathcal{G}=(\mathcal{G}_1,\cdots,\mathcal{G}_P)^T$. For simplicity, the dependence of $\bm{\widetilde{\beta}}$ on $u$ is omitted for simplicity.

{\bf Training Loss.} 
For simplicity, we denote both the time differentiation operator and the Laplace operator as:
\begin{equation}
	\mathcal{R}^n(u) = \frac{\bm{\beta}^n(u)-\bm{\beta}^{n-1}(u) }{t_n-t_{n-1}} + D\Lambda_P \bm{\beta}^n(u).
\end{equation}
To train the neural network $\mathcal{G}(\cdot,\theta):\mathbb{R}^P\rightarrow\mathbb{R}^P$, we focus on both the data RMSE loss and the model loss:
\begin{itemize}
	\item {\textbf{Data $L^2$ loss}
		ensures that the discretized solution $\bm{\widetilde{\beta}}^n$ closely matches the given data $\bm{\beta}^n$     
	}:
	\begin{equation}
		L^D(\theta):= \frac{1}{N}\sum_{n=1}^{N} \frac{\|\bm{\widetilde{\beta}}^n-\bm{\beta}^n\|_{l^2}}{\|\bm{\beta}^n\|_{l^2}}.
	\end{equation}
	\item 	{\textbf{Model residual loss}  guarantees that the solution satisfies the model problem}:
	\begin{equation}
		L^R(\theta):=\frac{1}{N}\sum_{n=1}^{N} \frac{ \|\mathcal{R}^n-\mathcal{G}(\bm{{\beta}}^{n-1};\theta)\|_{l^2}}{\|\mathcal{R}^n\|_{l^2}}.
	\end{equation}
\end{itemize}

Then we train the neural operator by minimizing the following combined loss function:
\begin{equation}
	\min_\theta L(\theta):= L^D(\theta)+L^R(\theta).
\end{equation}



\subsection{Incorporating $A$–$\tau$–$N$ Pathology}

To incorporate the $A$–$\tau$–$N$ pathology consistent with the amyloid cascade hypothesis \cite{jack2018nia,hardy2002amyloid}, we apply our framework using the eigenfunctions $\{\phi_i\}_{i=1}^P$ of a positive-definite operator $L$ as basis functions to construct neural operators for each pathological component. Specifically, we approximate the nonlinear terms $\mathcal{F}_1$, $\mathcal{F}_2$, and $\mathcal{F}_3$ as follows:
\begin{equation}
	\begin{aligned}
		\mathcal{F}_1(A) &\approx \mathcal{N}_1(A) = \sum_{i=1}^{P} \mathcal{G}^1_i(\bm\beta(A);\theta)\, \phi_i, \\
		\mathcal{F}_2(A, \tau) &\approx \mathcal{N}_2(A, \tau) = \sum_{i=1}^{P} \mathcal{G}^2_i(\bm\beta(A), \bm\beta(\tau);\theta)\, \phi_i, \\
		\mathcal{F}_3(A, \tau, N) &\approx \mathcal{N}_3(A, \tau, N) = \sum_{i=1}^{P} \mathcal{G}^3_i(\bm\beta(A), \bm\beta(\tau), \bm\beta(N);\theta)\, \phi_i,
	\end{aligned}
\end{equation}
where each $\mathcal{G}^j_i$ (for $j=1,2,3$ and $i=1,\ldots,P$) is a neural network mapping input states to spectral coefficients. The operators $\mathcal{N}_1$, $\mathcal{N}_2$, and $\mathcal{N}_3$ are trained sequentially following this structure. The detailed neural network architectures are provided in the \textbf{Appendix}.

To accommodate inter-subject variability in disease progression rates (e.g., due to age or disease stage) \cite{jedynak2012computational,zheng2022data}, we introduce a learnable temporal scaling factor $\xi$, such that time is rescaled as $s = \xi t$. For example, the evolution equation for $A$ becomes:
\[
	\xi u_t - \alpha_u \Delta u = \mathcal{F}(u),
\]
where $\xi$ is patient-specific and learned from longitudinal data.

We evaluate model performance across $M$ patient trajectories and $N$ time points per trajectory using three metrics:
\begin{itemize}
	\item \textbf{Relative $L^2$ error of the solution} ($E_{L^2}$):
	\[
		\frac{1}{MN} \sum_{m=1}^M \sum_{n=1}^{N} \frac{\| u^n_m - u_m(t_n) \|_{L^2}}{\| u_m(t_n) \|_{L^2}}
	\]
	\item \textbf{Relative PDE residual error} ($E_{\mathrm{Res}}$):
	\[
		\frac{1}{MN} \sum_{m=1}^M \sum_{n=1}^{N} \frac{\| \mathcal{G}(\bm\beta^{n-1}(\widetilde{u}^n_m); \theta) - \mathcal{R}^n(u_m) \|_{l^2}}{\| \mathcal{R}^n(u_m) \|_{l^2}}
	\]
	\item \textbf{Relative nonlinear term error} ($E_{\mathrm{Nonlinear}}$):
	\[
		\frac{1}{MN} \sum_{m=1}^M \sum_{n=1}^{N} \frac{\| \mathcal{N}(\widetilde{u}^n_m) - \mathcal{F}(u_m(t_n)) \|_{L^2}}{\| \mathcal{F}(u_m(t_n)) \|_{L^2}}
	\]
\end{itemize}
These metrics collectively assess the model's accuracy in approximating the solution, satisfying the governing PDE constraints, and capturing the nonlinear terms, as summarized in Table~\ref{tab:combined_accuracy}.

After training, the learned neural operators are used to predict future system dynamics across different stages of disease progression. To extract mechanistic insights, we compute the Jacobian of each neural operator at representative data points corresponding to clinical stages such as MCI and AD. This analysis reveals interpretable patterns of directional influence and quantifies effective interaction lengths between brain regions, shedding light on the underlying connectivity structure inferred from data.

\subsection{Learning Cognitive Score Dynamics}

For the cognitive score $C$ described in Equation~\eqref{eq:c}, the governing equation reduces to a scalar ordinary differential equation (ODE). To model this, we directly approximate its nonlinear term using a neural network:
\begin{equation}
\mathcal{F}_4\left( N \right) \approx \mathcal{N}_4(\bm\beta(N)),
\end{equation}
where the input is the regional neurodegeneration profile $N \in \mathbb{R}^{68}$ projected onto the eigenfunction space, and the output represents the predicted rate of cognitive decline. After training, the neural network $\mathcal{N}_4$ enables forecasting of future cognitive scores, conditioned on the dynamically evolving neurodegeneration patterns inferred from the $A$–$\tau$–$N$ model.

\subsection{Personalized Treatment Strategy}

Building upon the learned dynamical system in Equation~\eqref{eq:atauN}, we introduce personalized intervention terms, $d_A$ and $d_\tau$, to modulate the dynamics of anti-A$\beta$ and anti-tau therapies, respectively~\cite{van2023lecanemab,mintun2021donanemab,edland2023semorinemab,cummings2021alzheimer}. These treatment dosing profiles are parameterized as neural networks:
\begin{equation}
\begin{aligned}
d_A \approx \mathcal{N}_A(t), \quad
d_\tau \approx \mathcal{N}_\tau(t),
\end{aligned}
\end{equation}
where $t$ denotes patient age, and both $\mathcal{N}_A$ and $\mathcal{N}_\tau$ are individualized for each subject. To enforce biological feasibility, the outputs of these networks are constrained to be non-negative and bounded by applying a sigmoid activation function at their final layers. The detailed neural network architectures are provided in the \textbf{Appendix}.

Using these personalized treatment profiles, we simulate a range of intervention strategies—including anti-A$\beta$, anti-tau, and combination therapies—by incorporating the control terms directly into the system’s governing equations.

To identify optimal dosing schedules, we formulate an objective function that balances cognitive benefit against treatment burden:
\begin{equation}
\min_{d_A, d_\tau} \left( -C + \eta_A d_A^2 + \eta_\tau d_\tau^2 \right),
\end{equation}
where $C$ denotes the predicted cognitive score, and higher values correspond to better cognitive function. The negative sign in front of $C$ ensures that the optimization seeks to maximize cognitive outcomes. The regularization terms $\eta_A$ and $\eta_\tau$ penalize excessively high treatment intensities, preserving biological plausibility and limiting potential adverse effects. This formulation allows us to design individualized, precision intervention strategies tailored to each patient's disease trajectory, balancing therapeutic benefit with clinical feasibility.

\subsection{Data Availability}

\textbf{Synthetic Data on 2D Brain Geometry.}  We conducted experiments using synthetic data generated on an unstructured 2D brain-shaped domain. In this setting, the finite element method (FEM) was employed to numerically compute the eigenfunctions of the Laplacian operator, which served as the spectral basis for the neural operator. The model was trained on multiple initial conditions and age-dependent trajectories to learn the system dynamics. Once trained, the model was evaluated by predicting the system’s behavior at future, unobserved time points.

\textbf{ADNI Dataset}  
Data used in the preparation of this article were obtained from the ADNI database (\url{http://adni.loni.usc.edu/}). ADNI was launched in 2003 as a public-private partnership, led by Principal Investigator Michael W. Weiner, MD. Its primary goal is to determine whether serial magnetic resonance imaging (MRI), positron emission tomography (PET), other biological markers, and clinical and neuropsychological assessments can be combined to track the progression of mild cognitive impairment (MCI) and early Alzheimer’s disease (AD).

For this study, we accessed multimodal neuroimaging and biomarker data from the ADNI database under an approved data use application. A$\beta$-PET and tau-PET standardized uptake value ratios (SUVRs) were obtained from the datasets titled \textit{``UC Berkeley – Amyloid PET 6mm Res analysis [ADNI1, GO, 2, 3, 4]''} and \textit{``UC Berkeley – Tau PET PVC 6mm Res analysis [ADNI2, 3, 4]''}, respectively. Regional SUVRs were computed by dividing the standardized uptake values (SUVs) in target regions by the SUV of the whole cerebellum, chosen as the reference region due to its minimal nonspecific binding and stable tracer uptake across subjects.

Cortical thickness measurements were derived from the \textit{``UCSF–Cross-Sectional FreeSurfer (6.0) [ADNI3]''} and \textit{``UCSF–Cross-Sectional FreeSurfer (5.1) [ADNI1, GO, 2]''} datasets. 

Subjects were not required to have data from all imaging modalities. Instead, inclusion criteria specified that each subject must have data from at least three separate visits for one or more of the following modalities: tau-PET, A$\beta$-PET, or cortical thickness.

\section*{Acknowledgements}
JW and WH were supported by the National Institute of General Medical Sciences through grant 1R35GM146894. 

\section*{For the Alzheimer's Disease Neuroimaging Initiative}

Funding for data collection was funded by the Alzheimer’s Disease Neuroimaging Initiative (ADNI) (National Institutes of Health Grant U01 AG024904, Michael Weiner, PI) and DOD ADNI (Department of Defense award number W81XWH-12–2–0012). ADNI is funded by the National Institute on Aging, the National Institute of Biomedical Imaging and Bioengineering, and through contributions from the following: AbbVie, Alzheimer ’s Association; Alzheimer ’s Drug Discovery Foundation; Araclon Biotech; BioClinica, Inc.; Biogen; Bristol-Myers Squibb Company; CereSpir, Inc.; Cogstate; Eisai Inc.; Elan Pharmaceuticals, Inc.; Eli Lilly and Company; EuroImmun; F. Hoffmann-La Roche Ltd and its affiliated company Genentech, Inc.; Fujirebio; GE Healthcare; IXICO
Ltd.; Janssen Alzheimer Immunotherapy Research \& Development, LLC.; Johnson \& Johnson Pharmaceutical Research \& Development LLC.; Lumosity; Lundbeck; Merck \& Co., Inc.; Meso Scale Diagnostics, LLC.; NeuroRx Research; Neurotrack Technologies; Novartis Pharmaceuticals Corporation; Pfizer Inc.; Piramal Imaging; Servier; Takeda Pharmaceutical Company; and Transition Therapeutics. The Canadian Institutes of Health Research is providing funds to support ADNI
clinical sites in Canada. Private sector contributions are facilitated by the Foundation for the National Institutes of Health (www.fnih.org). The grantee organization is the Northern California Institute for Research and Education, and the study is coordinated by the Alzheimer’s Therapeutic Research Institute at the University of Southern California. ADNI data are disseminated by the Laboratory for Neuro Imaging at the University of Southern California. ADNI investigators contributed to the design and implementation of the ADNI database and/or provided data but did not participate in the analysis or writing of this report. A complete listing of ADNI investigators can be found at \url{http://adni.loni.usc.edu/wp-content/uploads/how_to_apply/ADNI_Acknowledgement_List.pdf}.

\bibliographystyle{plain}
\bibliography{main}

\begin{thebibliography}{10}

\bibitem{pmid36847013}
Craig~S Atwood and George Perry.
\newblock Playing russian roulette with {A}lzheimer’s disease patients: Do the cognitive benefits of lecanemab outweigh the risk of edema, stroke and encephalitis?
\newblock {\em Journal of Alzheimer’s Disease}, 92(3):799--801, 2023.

\bibitem{boffi2010finite}
Daniele Boffi.
\newblock Finite element approximation of eigenvalue problems.
\newblock {\em Acta numerica}, 19:1--120, 2010.

\bibitem{pmid36834612}
Francesco Bruno, Paolo Abondio, Alberto Montesanto, Donata Luiselli, Amalia~C Bruni, and Raffaele Maletta.
\newblock The nerve growth factor receptor (ngfr/p75ntr): A major player in {A}lzheimer’s disease.
\newblock {\em International journal of molecular sciences}, 24(4):3200, 2023.

\bibitem{cummings2021alzheimer}
Jeffrey Cummings, Garam Lee, Kate Zhong, Jorge Fonseca, and Kazem Taghva.
\newblock Alzheimer's disease drug development pipeline: 2021.
\newblock {\em Alzheimer's \& Dementia: Translational Research \& Clinical Interventions}, 7(1):e12179, 2021.

\bibitem{edland2023semorinemab}
Steven~D Edland and Jorge~J Llibre-Guerra.
\newblock Semorinemab in mild-to-moderate {A}lzheimer disease, 2023.

\bibitem{pmid36818565}
Ranya Elsayed, Mahmoud Elashiry, Yutao Liu, Ana~C Morandini, Ahmed El-Awady, Mohamed~M Elashiry, Mark Hamrick, and Christopher~W Cutler.
\newblock Microbially-induced exosomes from dendritic cells promote paracrine immune senescence: novel mechanism of bone degenerative disease in mice.
\newblock {\em Aging and Disease}, 14(1):136, 2023.

\bibitem{pmid20083042}
Mitchell~R Goldsworthy and Ann-Maree Vallence.
\newblock The role of $\beta$-amyloid in {A}lzheimer's disease-related neurodegeneration.
\newblock {\em Journal of Neuroscience}, 33(32):12910--12911, 2013.

\bibitem{gueorguieva2023disease}
Ivelina Gueorguieva, Laiyi Chua, Brian~A Willis, John~R Sims, and Alette~M Wessels.
\newblock Disease progression model using the integrated {A}lzheimer's disease rating scale.
\newblock {\em Alzheimer's \& Dementia}, 19(6):2253--2264, 2023.

\bibitem{pmid30863455}
Wenrui Hao, Suzanne Lenhart, and Jeffrey~R Petrella.
\newblock Optimal anti-amyloid-beta therapy for {A}lzheimer’s disease via a personalized mathematical model.
\newblock {\em PLoS computational biology}, 18(9):e1010481, 2022.

\bibitem{hao2025laplacian}
Wenrui Hao and Jindong Wang.
\newblock Laplacian eigenfunction-based neural operator for learning nonlinear partial differential equations.
\newblock {\em arXiv preprint arXiv:2502.05571}, 2025.

\bibitem{hardy2002amyloid}
John Hardy and Dennis~J Selkoe.
\newblock The amyloid hypothesis of {A}lzheimer's disease: progress and problems on the road to therapeutics.
\newblock {\em science}, 297(5580):353--356, 2002.

\bibitem{iturria2017multifactorial}
Yasser Iturria-Medina, F{\'e}lix~M Carbonell, Roberto~C Sotero, Francois Chouinard-Decorte, Alan~C Evans, {A}lzheimer's~{D}isease Neuroimaging~Initiative, et~al.
\newblock Multifactorial causal model of brain (dis) organization and therapeutic intervention: Application to {A}lzheimer’s disease.
\newblock {\em Neuroimage}, 152:60--77, 2017.

\bibitem{iturria2016early}
Yasser Iturria-Medina, Roberto~C Sotero, Paule~J Toussaint, Jos{\'e}~Mar{\'i}a Mateos-P{\'e}rez, and Alan~C Evans.
\newblock Early role of vascular dysregulation on late-onset {A}lzheimer’s disease based on multifactorial data-driven analysis.
\newblock {\em Nature communications}, 7(1):11934, 2016.

\bibitem{jack2018nia}
Clifford~R Jack~Jr, David~A Bennett, Kaj Blennow, Maria~C Carrillo, Billy Dunn, Samantha~Budd Haeberlein, David~M Holtzman, William Jagust, Frank Jessen, Jason Karlawish, et~al.
\newblock Nia-aa research framework: toward a biological definition of alzheimer's disease.
\newblock {\em Alzheimer's \& dementia}, 14(4):535--562, 2018.

\bibitem{jagust2018imaging}
William Jagust.
\newblock Imaging the evolution and pathophysiology of {A}lzheimer disease.
\newblock {\em Nature Reviews Neuroscience}, 19(11):687--700, 2018.

\bibitem{jedynak2012computational}
Bruno~M Jedynak, Andrew Lang, Bo~Liu, Elyse Katz, Yanwei Zhang, Bradley~T Wyman, David Raunig, C~Pierre Jedynak, Brian Caffo, Jerry~L Prince, et~al.
\newblock A computational neurodegenerative disease progression score: method and results with the {A}lzheimer's disease neuroimaging initiative cohort.
\newblock {\em Neuroimage}, 63(3):1478--1486, 2012.

\bibitem{li2018bayesian}
Dan Li, Samuel Iddi, Wesley~K Thompson, Michael~S Rafii, Paul~S Aisen, Michael~C Donohue, and Alzheimer's Disease~Neuroimaging Initiative.
\newblock Bayesian latent time joint mixed-effects model of progression in the {A}lzheimer's disease neuroimaging initiative.
\newblock {\em Alzheimer's \& Dementia: Diagnosis, Assessment \& Disease Monitoring}, 10(1):657--668, 2018.

\bibitem{pmid36807325}
Samantha~M Loi, Monica Cations, and Dennis Velakoulis.
\newblock Young-onset dementia diagnosis, management and care: a narrative review.
\newblock {\em Medical Journal of Australia}, 218(4):182--189, 2023.

\bibitem{michno2022chemical}
Wojciech Michno, Srinivas Koutarapu, Rafael Camacho, Christina Toomey, Katie Stringer, Karolina Minta, Junyue Ge, Durga Jha, Julia Fernandez-Rodriguez, Gunnar Brinkmalm, et~al.
\newblock {Chemical traits of cerebral amyloid angiopathy in familial British-, Danish-, and NON-{A}LZHEIMER's dementias}.
\newblock {\em Journal of Neurochemistry}, 163(3):233--246, 2022.

\bibitem{mintun2021donanemab}
Mark~A Mintun, Albert~C Lo, Cynthia Duggan~Evans, Alette~M Wessels, Paul~A Ardayfio, Scott~W Andersen, Sergey Shcherbinin, JonDavid Sparks, John~R Sims, Miroslaw Brys, et~al.
\newblock Donanemab in early {A}lzheimer’s disease.
\newblock {\em New England Journal of Medicine}, 384(18):1691--1704, 2021.

\bibitem{moravveji2024scoping}
Seyedadel Moravveji, Nicolas Doyon, Javad Mashreghi, and Simon Duchesne.
\newblock A scoping review of mathematical models covering {A}lzheimer's disease progression.
\newblock {\em Frontiers in Neuroinformatics}, 18:1281656, 2024.

\bibitem{petrella2019computational}
Jeffrey~R Petrella, Wenrui Hao, Adithi Rao, and P~Murali Doraiswamy.
\newblock Computational causal modeling of the dynamic biomarker cascade in {A}lzheimer’s disease.
\newblock {\em Computational and mathematical methods in medicine}, 2019(1):6216530, 2019.

\bibitem{rabiei2025data}
Kobra Rabiei, Jeffrey~R Petrella, Suzanne Lenhart, Chun Liu, P~Murali Doraiswamy, and Wenrui Hao.
\newblock Data-driven modeling of amyloid-beta targeted antibodies for {A}lzheimer's disease.
\newblock {\em arXiv preprint arXiv:2503.08938}, 2025.

\bibitem{raj2012network}
Ashish Raj, Amy Kuceyeski, and Michael Weiner.
\newblock A network diffusion model of disease progression in dementia.
\newblock {\em Neuron}, 73(6):1204--1215, 2012.

\bibitem{raket2020statistical}
Lars~Lau Raket.
\newblock Statistical disease progression modeling in {A}lzheimer disease.
\newblock {\em Frontiers in big Data}, 3:24, 2020.

\bibitem{pmid36804755}
Ovais Shafi, Uzair Yaqoob, Madiha Haseeb, Manwar Madhwani, Luqman~Naseer Virk, Syed Wajahat~Ali Zaidi, Ammar Nadeem, et~al.
\newblock Investigating the gliogenic genes and signaling pathways in the pathogenesis of huntington’s disease: A systematic review.
\newblock 2024.

\bibitem{susanne1994mathematical}
C~Susanne, L~Brenner, and LR~Scott.
\newblock The mathematical theory of finite element methods.
\newblock {\em Texts in Applied Mathematics}, 15, 1994.

\bibitem{pmid37183523}
Reina Tonegawa-Kuji, Yuan Hou, Bo~Hu, Noah Lorincz-Comi, Andrew~A Pieper, Babak Tousi, James~B Leverenz, and Feixiong Cheng.
\newblock Efficacy and safety of passive immunotherapies targeting amyloid beta in {A}lzheimer’s disease: A systematic review and meta-analysis.
\newblock {\em PLoS Medicine}, 22(3):e1004568, 2025.

\bibitem{pmid36872303}
Shivendra~Mani Tripathi, Sudhanshu Mishra, Rishabha Malviya, and Smriti Ojha.
\newblock Bioactive compounds targeting neurodegenerative diseases.
\newblock 2025.

\bibitem{van2023lecanemab}
Christopher~H Van~Dyck, Chad~J Swanson, Paul Aisen, Randall~J Bateman, Christopher Chen, Michelle Gee, Michio Kanekiyo, David Li, Larisa Reyderman, Sharon Cohen, et~al.
\newblock Lecanemab in early {A}lzheimer’s disease.
\newblock {\em New England Journal of Medicine}, 388(1):9--21, 2023.

\bibitem{pmid36829875}
David Vogrinc, Milica Gregori{\v{c}}~Kramberger, Andreja Emer{\v{s}}i{\v{c}}, Sa{\v{s}}a {\v{C}}u{\v{c}}nik, Katja Gori{\v{c}}ar, and Vita Dol{\v{z}}an.
\newblock Genetic polymorphisms in oxidative stress and inflammatory pathways as potential biomarkers in {A}lzheimer’s disease and dementia.
\newblock {\em Antioxidants}, 12(2):316, 2023.

\bibitem{zheng2022data}
Haoyang Zheng, Jeffrey~R Petrella, P~Murali Doraiswamy, Guang Lin, Wenrui Hao, and Alzheimer’s Disease~Neuroimaging Initiative.
\newblock Data-driven causal model discovery and personalized prediction in {A}lzheimer's disease.
\newblock {\em NPJ digital medicine}, 5(1):137, 2022.

\end{thebibliography}

\appendix
\section{Appendix: Training Details}

All models are trained using the Adam optimizer for 5000 epochs. The initial learning rate is set to $10^{-3}$ and reduced by a factor of 0.5 every 1000 epochs to promote convergence.

\subsection{Neural Network Architecture}

We employ fully connected neural networks to parameterize the neural operators $\mathcal{N}$. The spectral coefficient networks $\mathcal{G}_1$, $\mathcal{G}_2$, and $\mathcal{G}_3$ use the ReLU activation function. For the treatment control networks $\mathcal{N}_A$ and $\mathcal{N}_\tau$, sigmoid activations are used to enforce non-negativity and bounded outputs.

For the $A$--$\tau$--$N$ dynamics, we use 48 Laplacian eigenfunctions as the spectral basis. In simulations involving unstructured 2D brain geometries, 64 eigenfunctions are employed to better capture complex spatial patterns.

A summary of the neural network architectures is provided in Table~\ref{tab:arc}.

\begin{table}[!htbp]
    \centering
    \caption{Neural network architectures for different components of the framework.}
    \label{tab:arc}
    \begin{tabular}{c|cc}
        \hline
        \textbf{Neural Network} & \textbf{Activation Function} & \textbf{Layers} \\
        \hline
        \multicolumn{3}{c}{\textit{Equation~\eqref{eq:atauN}}} \\
        \hline
        $\mathcal{G}_1$ & ReLU & 48-128-128-48 \\
        $\mathcal{G}_2$ & ReLU & 96-128-128-48 \\
        $\mathcal{G}_3$ & ReLU & 144-128-128-48 \\
        $\mathcal{N}_4$ & ReLU & 48-128-128-1 \\
        $\mathcal{N}_A$ & Sigmoid & 1-128-128-1 \\
        $\mathcal{N}_\tau$ & Sigmoid & 1-128-128-1 \\
        \hline
        \multicolumn{3}{c}{\textit{Equation~\eqref{eq:abetanbrain}}} \\
        \hline
        $\mathcal{G}_1$ & ReLU & 64-100-100-100-64 \\
        $\mathcal{G}_2$ & ReLU & 128-100-100-100-64 \\
        $\mathcal{G}_3$ & ReLU & 192-100-100-100-64 \\
        $\mathcal{N}_4$ & ReLU & 64-100-100-100-1 \\
        \hline
    \end{tabular}
\end{table}

\subsection{Data and Parameter Settings}

The standardized uptake value ratio (SUVR) data used in this study are derived from PET imaging.

For the 2D brain-shaped simulations described in Equation~\eqref{eq:abetanbrain}, we generate synthetic spatiotemporal trajectories using a reaction–diffusion system with biologically inspired parameter values. The parameter settings are summarized in Table~\ref{tab:synthetic_params}.

\begin{table}[!htbp]
    \centering
    \caption{Parameter values for the reaction–diffusion system in Equation~\eqref{eq:abetanbrain}.}
    \label{tab:synthetic_params}
    \begin{tabular}{|c|c|}
        \hline
        \textbf{Parameter} & \textbf{Value} \\
        \hline
        $\alpha_A$ & 1 \\
        $\lambda_A$ & 0.4 \\
        $K_A$ & 1 \\
        $\alpha_\tau$ & 1 \\
        $\lambda_{\tau A}$ & 0.1 \\
        $\lambda_{\tau}$ & 0.2 \\
        $K_\tau$ & 1 \\
        $\alpha_N$ & 1 \\
        $\lambda_{N\tau}$ & 0.1 \\
        $\lambda_{N}$ & 0.2 \\
        $K_N$ & 1 \\
        $\lambda_{CN}$ & 0.005 \\
        $\lambda_{C}$ & 0.2 \\
        $K_{C}$ & 1 \\
        \hline
    \end{tabular}
\end{table}

\end{document}